\documentclass[lettersize,journal]{IEEEtran}
\usepackage{amsmath,amsfonts}
\usepackage{algorithmic}
\usepackage{algorithm}
\usepackage{array}
\usepackage[caption=false,font=footnotesize]{subfig}
\usepackage{textcomp}
\usepackage{stfloats}
\usepackage{url}
\usepackage{verbatim}
\usepackage{graphicx}
\usepackage{cite}
\usepackage{xcolor}
\usepackage{color}
\usepackage{multirow}

\newcolumntype{P}[1]{>{\centering\arraybackslash}p{#1}}
\definecolor{green(html/cssgreen)}{rgb}{0.0, 0.5, 0.0}

\newcommand{\rv}[1]{\textcolor{red}{#1}}
\newcommand{\ov}[1]{\textcolor{orange}{#1}}
\newcommand{\bv}[1]{\textcolor{blue}{#1}}
\newcommand{\gv}[1]{\textcolor{green(html/cssgreen)}{#1}}

\begin{document}

%\title{Exploring Large-Scale Data in \\ Fully-Unsupervised Re-Identification Setups}

\title{Large-scale Fully-Unsupervised Re-Identification}

%\author{IEEE Publication Technology,~\IEEEmembership{Staff,~IEEE,}

\author{Gabriel~Bertocco,
        Fernanda~Andal\'{o},~\IEEEmembership{Member,~IEEE,}
        Terrance~E.~Boult,~\IEEEmembership{Fellow~Member,~IEEE,} and~Anderson~Rocha,~\IEEEmembership{Senior~Member,~IEEE}% <-this % stops a space
\thanks{Gabriel Bertocco is a Ph.D. student at the Artificial Intelligence Lab. (\textbf{Recod.ai}), Institute of Computing, University of Campinas, Brazil}% <-this % stops a space
\thanks{Fernanda~Andal\'{o} is a researcher associated with the Artificial Intelligence Lab. (\textbf{Recod.ai}), Institute of Computing, University of Campinas, Brazil.}% <-this % stops a space
\thanks{Terrance~Boult is a professor at University of Colorado Colorado Springs (UCCS) and Chair of the Vision And Security Technology Lab (UCCS VAST LAB).}% <-this %
\thanks{Anderson Rocha is a Full-Professor for Artificial Intelligence and Digital Forensics and Chair of the Artificial Intelligence Lab. (\textbf{Recod.ai}) at the Institute of Computing, University of Campinas, Brazil. 
}}% <-this % stops a space

\markboth{Paper submitted for a possible publication in an IEEE Transactions}%
{Bertocco \MakeLowercase{\textit{et al.}}: Large-scale Fully-Unsupervised Re-Identification}


\maketitle

\begin{abstract}
Fully-unsupervised Person and Vehicle Re-Identification have received increasing attention due to their broad applicability in areas such as surveillance, forensics, event understanding, and smart cities, without requiring any manual annotation. However, most of the prior art has been evaluated in datasets that have just a couple thousand samples. Such small-data setups often allow the use of costly techniques in terms of time and memory footprints, such as Re-Ranking, to improve clustering results. Moreover, some previous work even pre-selects the best clustering hyper-parameters for each dataset, which is unrealistic in a large-scale fully-unsupervised scenario. In this context, this work tackles a more realistic scenario and proposes two strategies to learn from large-scale unlabeled data. The first strategy performs a local neighborhood sampling to reduce the dataset size in each iteration without violating neighborhood relationships. A second strategy leverages a novel Re-Ranking technique, which has a lower time upper bound complexity and reduces the memory complexity from $\mathcal{O}(n^{2})$ to $\mathcal{O}(kn)$ with $k \ll n$. To avoid the need for pre-selection of specific hyper-parameter values for the clustering algorithm, we also present a novel scheduling algorithm that adjusts the density parameter during training, to leverage the diversity of samples and keep the learning robust to noisy labeling. Finally, due to the complementary knowledge learned by different models in an ensemble, we also introduce a co-training strategy that relies upon the permutation of predicted pseudo-labels, among the backbones, with no need for any hyper-parameters or weighting optimization. The proposed methodology outperforms the state-of-the-art methods in well-known benchmarks and in the challenging large-scale Veri-Wild dataset, with a faster and memory-efficient Re-Ranking strategy, and a large-scale, noisy-robust, and ensemble-based learning approach.    
\end{abstract}


\section{Introduction}
\label{sec:introduction}

\IEEEPARstart{C}{lustering}-based self-supervised learning has attracted attention in the last years due to its capability of learning from fully-unlabeled datasets by performing clustering and finetuning with pseudo-labels. However, most works, like SWaV~\cite{caron2020unsupervised} and DeepCluster~\cite{caron2018deep}, employ a clustering-based strategy as a pretext task and further finetune the model in the labeled (or partially labeled) downstream task such as image classification or semantic segmentation. In this context, those models leverage the self-supervised solution as an intermediate state before performing labeled fine-tuning, instead of directly applying them to some task. Other works, such as ASPC-DA~\cite{guo2019adaptive}, employ the clustering-based solution directly to the target task without requiring a pretext task and, consequently, no labeled data. However, most of them have focused on simpler datasets with a lower number of classes, such as MNIST~\cite{lecun1998gradient} and FASHION~\cite{xiao2017fashion}. %, and USPS\footnote{http://www.cad.zju.edu.cn/home/dengcai/Data/MLData.html}. 

Conversely, we leverage clustering-based self-supervised learning as the final solution for deployment, but in more complex scenarios, with challenging intra-class variations and a greater number of classes in large-scale scenarios. Specifically, we focus on fully-unsupervised Person and Vehicle Re-Identification. 
Since it is a fully-unsupervised problem, we do not consider any task-related information or labeling to design our solution, opening the path for further applications beyond re-identification tasks.

%\IEEEPARstart{F}{ully}-unsupervised Person and Vehicle Re-
The fully-unsupervised Person and Vehicle Re-
Identification tasks have a myriad of applications, such as in surveillance, forensics, event understanding, intelligence monitoring, and smart cities, without requiring any label or side information for training. Moreover, with the growing amount of data recorded daily from a range of heterogeneous sources, from surveillance cameras to aerial and mobile devices (e.g., UAVs), the fully-unsupervised re-identification task can face applications with hundreds of thousands of unlabeled samples. The development of fully-unsupervised and scalable re-identification methods is then paramount to solving these more demanding and realistic scenarios.

One obstacle to the deployment of re-identification state-of-art methods is that most of them have reached top-tier performance in small datasets. The largest well-known Person Re-Identification benchmarks \texttt{Market1501}~\cite{zheng2015scalable}, \texttt{DukeMTMC-ReID}~\cite{ristani2016performance}\footnote{\texttt{DukeMTMC-ReID} has been discontinued, and it must not be used for evaluation and benchmarking anymore. For this reason and following the recent literature, we \textbf{do not} use it for evaluation. More details in \textit{https://www.dukechronicle.com/article/2019/06/duke-university-facial-
recognition-data-set-study-surveillance-video-students-china-uyghur}}, and \texttt{MSMT17}~\cite{wei2018person} have 12,936, 16,522, and 32,621 training images, respectively. For Vehicle Re-Identification, the \texttt{Veri}~\cite{liu2016large} dataset has 37,778 training images. Since the training sets are rather small, researchers could employ effective but costly techniques to achieve top-tier performance, such as Re-Ranking~\cite{zhong2017re} and Co-Training~\cite{ge2020mutual, zhai2020multiple, zhai2022population}. The Re-Ranking technique~\cite{zhong2017re} has memory complexity of $\mathcal{O}(n^2)$ (where $n$ is the training set size) and time complexity of $\mathcal{O}(n^3 + nk^3)$ (where $k$ is the number of reciprocal neighbors, see Section~\ref{subsec:local_reranking_clustering} for further details). And Co-Training usually involves cross-supervision where the confidence level of the samples from one model is used to weigh the loss functions from the other models in the ensemble, and all loss functions are optimized at once with many hyper-parameters to control the contribution for each term. Furthermore, these hyper-parameter values can be hard to tune, mainly in large-scale fully-unsupervised datasets without a validation set. 

Aiming to deploy scalable and affordable solutions for large-scale Unsupervised Person/Vehicle Re-Identification, we design a novel pipeline to alleviate the discussed challenges. We propose a novel Re-Ranking method that leverages the previously proposed k-Reciprocal Encoding~\cite{zhong2017re} but considers just a local neighborhood to calculate the Jaccard distance, without relying on their set expansion nor in local query expansion. This allows us to reduce the time complexity and keep the memory complexity in $\mathcal{O}(kn)$ where ($k \ll n$). 

Moreover, we also propose a sampling method based on the local neighborhood for a randomly chosen point. We have different local neighborhoods in each epoch, which effectively reduces the training set size without violating the neighborhood properties of the selected point. This preserves the hard-positive and hard-negative samples for effective training and, at the same time, reduces memory and time complexities. 

Since co-training has shown impressive performance in dealing with noisy data, we also propose a simple co-training strategy based on the co-training theory but not requiring any human supervision or hyper-parameter tuning. During training, we generate pseudo-labels for unlabeled training data using different convolutional neural network backbones. Each backbone generates its own feature space and then its own set of pseudo-labels. We propose switching the pseudo-labels among the backbones; thus, one backbone supervises the other through pseudo-label predictions. % To verify our performances, we evaluate our methods not just in the well-known benchmarks but also in the large-scale VehicleID~\cite{liu2016deep} and Veri-Wild~\cite{lou2019veri} vehicle datasets which have 113,346 and 277,797 training images which are $3\times$ and $7.35\times$ respectively bigger than the Veri training set (the biggest one among the well-known benchmarks). 

Beyond the discussed large-scale challenges, a fundamental aspect has been overlooked: the choice of the clustering hyper-parameter. Most methods employ DBSCAN~\cite{ester1996density}, which is controlled by the density parameter $\varepsilon$, and some works consider an optimal value for each dataset. For instance, AdaMG~\cite{peng2023adaptive} sets $\varepsilon = 0.5$ for \texttt{Market1501} and $\varepsilon = 0.7$ for \texttt{MSMT17}. When they fix the same value for both datasets, the performance drops. The recent ISE~\cite{zhang2022implicit} and RTMem~\cite{yin2023real} also set different hyper-parameter per dataset to achieve their best performance. However, since there is no validation set in a fully-unsupervised scenario, we argue it is unrealistic to select a specific value per dataset.

To alleviate the burden of selecting an optimal hyper-parameter per dataset, we analyze the clustering problem from the perspective of noisy-labeling-robust learning. During training, feature vectors are extracted, clustered, and pseudo-labels are assigned to them, which are then used for finetuning. In the first training epochs, the backbones have little knowledge about the dataset, so it is expected that the features are not too discriminative.
Therefore, there is more noisy pseudo labeling in the first iterations, which can require a tighter density parameter $\varepsilon$. As the model gets more robust during training, generating better features, it allows the loosening of the density parameter to include more hard-positive samples in the clusters. However, if we keep it loose until the end of the training, it might include too many non-matching samples, so we should decrease $\varepsilon$ after the feature space is reasonably tuned to allow final feature learning. Finally, if variations harden defining a stopping criterium, it might be interesting to hold the parameter constant at the end. This motivates us to design a new scheduling scheme for $\varepsilon$, which is used for all datasets without any hyperparameter tuning. We show that it reaches state-of-the-art performance in all evaluation scenarios, even outperforming the ones that select a dataset-specific hyper-parameter for optimal performance.
% where it starts at a low value, reaches a maximum in the middle of training, and decreases back until it reaches a plateau. We keep the same scheduling for all datasets without any hyperparameter tuning. We also show that it reaches state-of-the-art performance in all evaluation scenarios, even surpassing the ones that select a dataset-specific hyper-parameter for optimal performance.

To verify the performance of the proposed methods, we evaluate them not just in the well-known benchmarks but also in the large-scale \texttt{VehicleID}~\cite{liu2016deep} and \texttt{Veri-Wild}~\cite{lou2019veri} datasets, which are % have 113,346 and 277,797 training images which are 
$3\times$ and $7.35\times$ larger, respectively, than the Veri dataset (the largest among well-known benchmarks).

The key contributions of our work are:
\begin{itemize}
    \item A neighborhood-based sampling method to decrease the dataset size in each epoch. By preserving the neighborhood, it is able to keep hard-positive and hard-negative samples for model learning.  
    \item A Re-Ranking method that considers just the top-$k$ nearest neighbors and does not need set expansion nor local query expansion as previous methods. In this way, we effectively reduce the time and memory complexities. 
    \item A density parameter scheduling to deal with noisy data during training and, at the same time, bring diverse samples together. We keep the scheduling scheme for all datasets, avoiding hyper-parameter tuning.
    \item A co-training method where we switch the predicted labels among the involved backbones. This allows us to consider co-training without human intervention and parameter tuning. 
    \item We consider the large-scale \texttt{VehicleID} and \texttt{Veri-Wild datasets}, infrequently used in the prior art, to verify the model's performance in truly large-scale scenarios.
    
\end{itemize}

\section{Related Work}
\label{sec:related_work}

\subsection{Re-Ranking-based approaches}

Most Re-Ranking techniques are designed to address rank retrieval during the evaluation phase. Given a query image and the set of gallery images, the initial ranking list is obtained.
%which is the sorted list of the gallery images based on the similarity (or distance) to the query image. 
Usually, methods take this initial list and rank the samples again based on some strategy to enhance the retrieval. 

In~\cite{garcia2017discriminant}, the authors define the content and context sets. Given a probe image, they calculate its ranking list with the gallery images. The set with the closest images to the probe is called the content set, which is used with the original ranking list to create the context set and improve the ranking performance. 
%, and, based on the dissimilarity values, they divide the gallery images into three disjoint sets. The set with the closest images to the probe is called the content set, which is used along with the original probe ranking list to create the context set and improve the ranking performance.
In~\cite{bai2017scalable}, the authors use a graph-based on-the-fly affinity learning considering the labeled training, gallery, and probe sets to refine the ranking of the probe to the gallery set. In~\cite{sarfraz2018pose}, the authors employ the probe-to-neighbor distance and neighbor-to-neighbor distance to refine the ranking list by considering an expanded neighborhood from each gallery sample retrieved in the top matches to the probe. 
%They also employed the ranking-list distance based on~\cite{jarvis1973clustering} to avoid costly distance computation among different ranking lists. 
The authors in~\cite{luo2019spectral} propose the Spectral Feature Transformation (SFT), where they optimize the model to generate feature representation that optimizes the Min-Cut problem considering the labeled samples at batch level. During the evaluation, given a query sample, they perform SFT in the top-ranked gallery features to improve the retrieval performance.

The most well-known Re-Ranking method is the k-Reciprocal Encoding~\cite{zhong2017re} which has been extensively used for unsupervised re-identification. It improves the feature distances during training and generates clusters with high diversity and true-positive rate. The $k$-Reciprocal nearest neighbors are calculated for each training sample and respective expanded set. Then Jaccard distances are computed between training points considering each expanded set, and these distances are averaged in the Local Query Expansion step. %This process effectively increases clustering performance by including diverse examples in the clusters. 
However, it is time- and memory-consuming~\cite{jarvis1973clustering, luo2019spectral}. 

%They calculate the $k$-Reciprocal nearest neighbors $R(p,k)$ for each sample $p$ in the training set, and the set $R(q,k/2)$ where $q \in R(p,k)$. If $|R(p,k) \cap R(q,k/2)| \geq \frac{2}{3}|R(q,k/2)|$, they generate the expansion set $R*(p,k) = R(p,k) \cup R(q,k/2)$ to include the hard-positive samples. Then they calculate the Jaccard distances between training points considering each $R*(p,k)$ and average the distances in the Local Query Expansion step. This process effectively increases the clustering performance by including diverse examples on a cluster. However, it is time and memory-consuming~\cite{jarvis1973clustering, luo2019spectral}. 

%Since most of the Unsupervised Re-Identification works have employed their methods in relatively small datasets, such as MSMT17 and Veri, the time and memory costs have been affordable so far. However, as the number of training samples grows to more than 100K samples (e.g., VehicleID and Veri-Wild),
%the cost of employing Re-Ranking~\cite{zhong2017re} increases, making the training slower and with a high memory footprint.

In this context, we propose Local Re-Ranking, which redesigns the neighborhood-based distance calculation to decrease time complexity and keep state-of-the-art performance.
%We keep or even improve the state-of-art performance in the well-known Person and Vehicle benchmarks as well as in the large-scale VehicleID and Veri-Wild datasets. %with a less complex Re-Ranking technique. 

\subsection{Noise-robust Feature Learning}
\label{sec:noise_robust_learning}
% The problem of learning under noisy labels has received increasing attention since the publication of one of its seminal works~\cite {angluin1988learning}. 
Large-scale datasets are prone to annotation error due to their size or to the complexity in identifying the positive samples even with human supervision~\cite{frenay2013classification}. This introduces noise in the learning process, hindering generalization.

Several works have tackled model learning with noisy labels~\cite{frenay2013classification, han2018co, han2020survey, song2022learning}. They usually estimate the Transition Matrix, which reflects the probability of the samples from one class being misclassified as other classes present in the dataset. With the estimated Transition Matrix, some works~\cite{sukhbaatar2014training} propose to change the probability distribution on the final softmax layer or to reweight the final loss function~\cite{patrini2017making} to alleviate the influence of noisy samples. Based on the memorization effect in which Deep Neural Networks first learn from clean samples and then from noisy samples~\cite{han2018co, yu2019does, wang2019co}, some works have proposed robust regularization through implicit or explicit regularization~\cite{goodfellow2014explaining, xia2020robust}
and co-training methods~\cite{han2018co, wang2019co}. The fundamental idea of co-training in noisy-label scenarios is to have two (or more) backbones and select the small-loss samples from one peer backbone to train the other. Usually, the same batch of images is fed to both peers, and just the top-$r$\% small-loss samples are kept from each one. Then one peer optimizes its weights with the small-loss samples from another. Rate $r$ is decreased along the training to keep fewer samples to train the peers as they start learning from noisy samples due to the memorization effect. %For further details, we refer to the surveys~\cite{frenay2013classification, han2020survey, song2022learning}.

Based on the memorization effect, we propose to control the tightness of the generated clusters by changing the density criteria in a novel manner. But instead of selecting small-loss samples as done previously, we consider the presence of noisy labeling in the feature space.

\subsection{Co-training for Person Re-Identification}
\label{sec:co_training_reid}
Despite the progress in noise-robust feature learning, most methods assume data %(despite being noisy) 
is annotated. Instead, we consider unlabeled data and propose a clustering-based solution to generate pseudo-labels for model fine-tuning. Since the backbones have not been pre-trained in any other ReID-related dataset (just in ImageNet), the features are naturally noisy, which results in noisy labels and imperfect clustering. 

Prior Unsupervised Person ReID (UPReID) art also faces the same problem, and some works apply co-training to deal with noisy labeling. MMT~\cite{ge2020mutual}, MEB-Net~\cite{zhai2020multiple}, and PEG~\cite{zhai2022population} share the same principle of multiple models learning from each other's hard and soft labels at the batch level to encourage knowledge decoupling and robustness to noise. %They vary in the number of peer networks and in the strategy to train the models. MMT has two peer networks, while MEB has three and a regularization authority term to give more confidence to the peer that better groups the data in the clustering stage. PEG has a pool of backbones available, which are selected in turn to create the committee of peer networks. They also propose to change their hyper-parameters to imitate mutation to leverage an evolutionary-based approach. However, 
These methods employ complex co-training strategies with a lot of terms and hyper-parameters in their loss functions or in the selection of the peer networks, which might be challenging to tune and deploy in large-scale unlabeled scenarios. 

We propose a simpler co-training strategy, by permuting the predicted labels among the peers instead of performing soft/hard cross-supervision. We take advantage of co-training, but with a hyper-parameter-free strategy that does not require any manual or grid-searching-based parameter selection. 

\subsection{Unsupervised Re-Identification}

To tackle unsupervised re-identification, some methods rely on pre-training using a source ReID dataset~\cite{% zhong2020learning, zou2020joint, 
chen2020enhancing, 9521886, ge2020mutual, zhai2020multiple} to acquire prior knowledge. Other methods, like IICS~\cite{xuan2021intra}, CAP~\cite{wang2021camera}, CASTOR~\cite{xu2022pseudo}, and PPSL~\cite{wu2022pseudo},  rely on other information, such as camera labels. Since our method is fully unsupervised, i.e., it \textbf{does not} rely on camera labels, we focus on methods operating in the same setup. We present in the Supplementary Material a comparison table to methods considering camera labels as side information. 

%IICS~\cite{xuan2021intra} leverages intra-camera training by dividing samples into sets according to their camera labels and clustering each one. A backbone is trained in a multi-task manner (one task per camera), and clustering is run for the whole dataset, grouping samples of the same identity seen from different cameras.
%CAP~\cite{wang2021camera} performs clustering by assigning pseudo-labels for each sample of the dataset. They obtain camera proxies on each cluster for intra- and inter-camera training. 
ICE~\cite{Chen_2021_ICCV} has two versions: camera-aware and camera-agnostic. The first considers camera proxies for each cluster. The second considers only a cluster proxy, which is a feature average regardless of camera labels. They use a proxy-based loss, and hard- and soft-instance losses.
%CASTOR~\cite{xu2022pseudo} performs pre- and post-clustering processing to refine the distance matrix and pseudo-labels. In the pre-clustering, they leverage camera feature distances to regularize identity feature distances and avoid the cross-camera impact. In post-clustering, they proposed a $k$-reciprocal nearest neighbors-based search for outlier reassignment. Liu~\textit{et. al.}~\cite{liu2022unsupervised} also proposed to calculate the similarity between different camera domains and regularize the distances to consider cross-camera variance. They proposed a combination of instance- and cluster-based feature memories for model optimization. 
%PPSL~\cite{wu2022pseudo} performs patch-based clustering where surrogate vectors are generated for each class based on the extracted patches, and images from different cameras and the respective gradients are used to perform clustering. Moreover, it also considers intra-similarity learning to explore different parts of the same image, and inter-similarity learning to align features of same-cluster images across cameras.   
SpCL~\cite{ge2020self} uses a self-paced strategy that introduces some metrics to measure cluster reliability: cluster independence and compactness. If both are higher than predefined thresholds, the cluster is kept within the feature space. RLCC~\cite{zhang2021refining} refines clusters by a consensus among iterations. Pseudo-labels are created by considering the ones generated in previous iterations, keeping the training stable.
CACL~\cite{li2022cluster} proposes to suppress the dominant colors in images, providing a more robust feature description, and a novel pseudo-label refinement method.
CCons~\cite{dai2022cluster} uses contrastive learning, with which they select the hardest cluster sample in the batch to update the cluster centroid.
ISE~\cite{zhang2022implicit} synthesizes novel feature examples from real ones to refine sample distribution, aiming to generate clusters with a higher true positive rate, as well as avoiding subdivision of the samples from the same class in different clusters. 
PPLR~\cite{cho2022part} employs a part-based model that creates feature spaces from different parts of the feature map. In each one, they calculate the nearest neighbors of the samples and propose a cross-agreement metric to refine the proposed pseudo-labels.
GRACL~\cite{zhang2022global} keeps two memory banks where one holds the sample, from each cluster, most dissimilar from other clusters, and the other holds the sample most dissimilar from the positive samples. This approach encourages compactness and separability.
AdaMG~\cite{peng2023adaptive} performs clustering with different density parameters to generate multiple pseudo-labels. Then a teacher-student model is trained considering each set of pseudo-label, and the cluster feature memory bank is updated based on sample reliability, alleviating noisy labeling impact.   

For Unsupervised Vehicle Re-Identification (UVReID), MLPL~\cite{he2022multi} adopts a multi-level feature description by extracting a global feature and four local features for clustering and feature learning. They also propose converging and promoting stages to learn from global and local features separately and jointly, based on the consistency score of the global- and local-based clustering results. 
%VR-PROUD~\cite{bashir2019vr} trains a side network to predict vehicle colors in each cluster. Since the same vehicle is supposed to have the same color across all cameras, they get the dominant color in each cluster and discard the vehicles with different colors assuming those images are false-positive samples. 
RLCC~\cite{zhang2021refining} and PPLR~\cite{cho2022part} are also evaluated under the UVReID scenario.

%Compared to such methods, our formulation herein does not rely on any camera labeling, requiring only the bounding boxes of the objects of interest (e.g., people, vehicles); therefore, it best fits this last category of methods. Moreover, 
Our method differs from the others by proposing a regime to adapt a noise-robust density parameter over time, alleviating the burden of choosing optimal hyper-parameters, and a co-training strategy to avoid error amplification by the backbones.

\section{Proposed Method}
\label{sec:proposed_method}

\begin{figure*}[ht]
\centering
\includegraphics[width=6.7in]{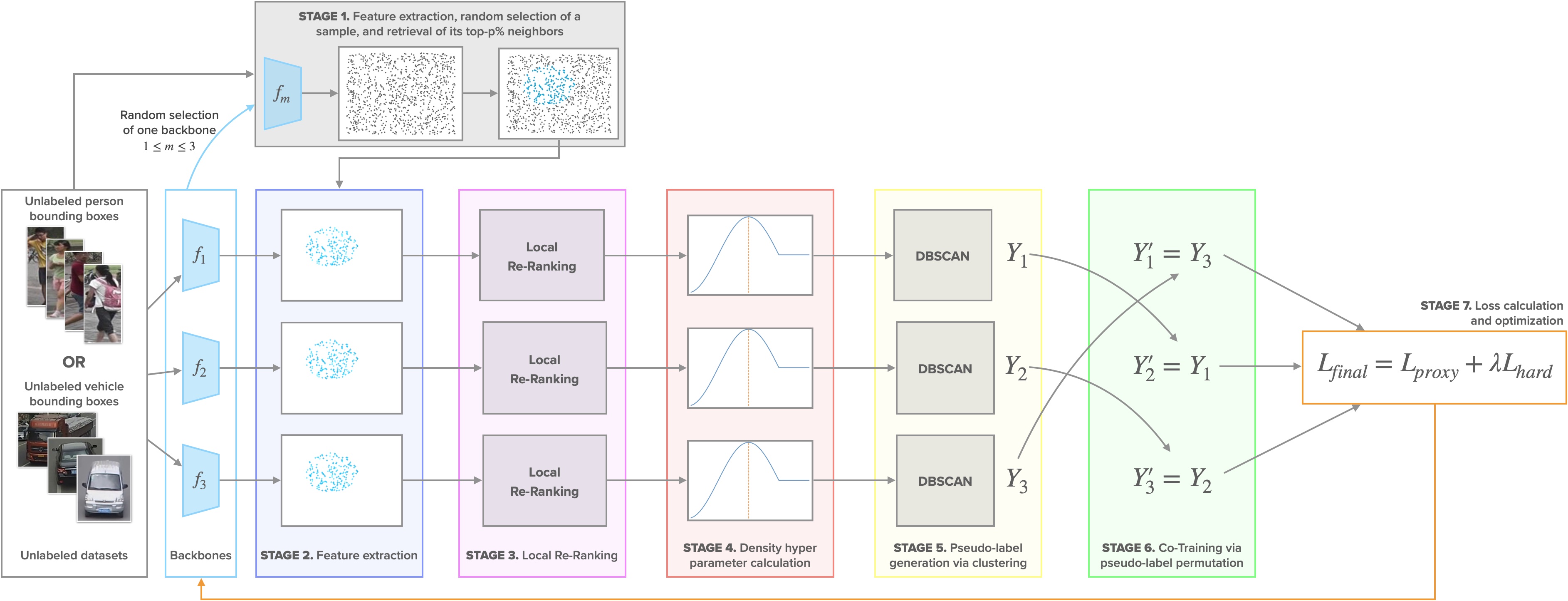}
\caption{Overview of our solution. In Stage 1, we first extract features for the entire unlabeled training set utilizing one randomly selected backbone. Then we randomly select a sample and obtain the top-$p\%$ closest neighbors to define the local neighborhood. In Stage 2, features are extracted for the selected samples, with all backbones. In Stage 3, we employ Local Re-Ranking to refine the distances based on the local neighborhood of the samples, keeping a low memory and time footprint. %than the previously~\cite{zhong2017re}. 
In Stage 4, we select the current density parameter $\varepsilon$ based on the proposed noise-robust density scheduling scheme. Then clustering is performed in Stage 5. Once we have the pseudo-labels predicted by each backbone in the pipeline, we permute the pseudo-labels set among the backbones to allow cross-supervision (Stage 6). The illustrated permutation ($Y_{1}^{'} = Y_{3}$, $Y_{2}^{'} = Y_{1}$, $Y_{3}^{'} = Y_{2}$) is an example, and other permutations are possible as long as a backbone is not supervised by its own pseudo-labels. % do not supervised its own backbone (i.e., $Y_{m}^{'} \neq Y_{m}$ for $1 \leq m \leq 3$). 
Finally, in Stage 7, the loss function is optimized. %See Section~\ref{sec:proposed_method} for a detailed explanation of each Stage. 
Best viewed in color.}
\label{fig:full_pipeline}
\end{figure*}

In this section, we provide the rationale for each part of the proposed pipeline (Figure~\ref{fig:full_pipeline}) to tackle large-scale unlabeled Re-Identification under noisy labeling. We assume we have $m = 3$ backbones, but it can be extended to any $m \geq 2$.

\subsection{Self-supervised Initialization}

Usually, previous methods adopt ResNet50~\cite{he2016deep} pre-trained on ImageNet~\cite{deng2009imagenet} for model initialization. Despite ImageNet being a general dataset for classification, the pre-trained backbone might not produce discriminant features for clustering in the UPReID and UVReID scenarios. More recently, self-supervised pretext tasks have been widely explored for initialization~\cite{caron2020unsupervised, he2020momentum, chen2020simple, caron2021emerging, zbontar2021barlow} and further application in downstream tasks.  In this context, we propose pre-training the model on the target unlabeled dataset before 
applying our pipeline. Since we assume a fully-unlabeled scenario, we 
leverage a self-supervised pre-training based on Barlow 
Twins~\cite{zbontar2021barlow}. %\rv{It does not rely on labels, and it is a negative-free mining approach since it relies only on the image and its respective augmentation without contrasting to the other images in the batch or dictionary.} 
We perform this initialization step for all backbones. More details are described in the Supplementary Material.

\subsection{Local Neighborhood Sampling (LNS)}
\label{sec:local_neighborhood_sampling}
It is a common requirement that the entire distance matrix, with all training feature vectors, be loaded into memory. However, as the number of considered samples grows, memory and time complexity increase which might lead to a costly deployment in a large-scale scenario. For instance, in~\cite{huang2019ultra}, the authors reported that a distance matrix of size $n \times n$, with $n = 10^{6}$, takes up $7450.58$ GB of memory. %, assuming each matrix entry is in double-precision format.
The largest datasets used in prior Unsupervised ReID works are \texttt{MSMT17} for Person and \texttt{Veri} for Vehicle Re-Identification. Since they have fewer than $40K$ training samples, previous methods might not have faced significant memory and time issues. However, there are several practical scenarios in which larger datasets are required~\cite{padilha2020forensic}. In these cases, a more efficient solution is required. 

To tackle this, given an unlabeled dataset $X = \{x_{i}^{n}\}$ with $n$ images, we propose a novel neighborhood-based approach for sampling a subset $X_{s}\subset X$ (Stage 1, Figure~\ref{fig:full_pipeline}). More specifically, we first select a backbone $f_{m}$ ($1 \leq m \leq 3$) randomly from the available backbones to perform feature extraction and obtain the feature vector set $F_{m}$ (blue points in Stage 1, Figure~\ref{fig:full_pipeline}). Then we randomly select a feature vector $v \in F_{m}$ and calculate its top-$p\%$ ($0 \leq p \leq 100$) nearest neighbors set in $F_{m}$, which defines $X_{s}\subset X$ (yellow points in Stage 1, Figure~\ref{fig:full_pipeline}). That is, $X_{s}$ comprises the elements of the whole dataset that are closer to $v$. The computational cost is related to computing the distances from $v$ to every other vector in $F_{m}$ and sorting the distance vector. Since cosine distance is used, the complexity is $\mathcal{O}(nd + nlgn)$, where $d$ is the feature dimension. In a practical scenario, we usually have $k \ll n$, and the distance computation is performed in parallel, so the actual complexity is $\mathcal{O}(nlgn)$. Once the set $X_{s}$ is defined, it is used by all backbones to perform feature extraction and Re-Ranking in Stages 2 and 3, respectively (blue points in Figure~\ref{fig:full_pipeline}). The set $X_{s}$ is redefined every three pipeline iterations by another randomly selected backbone, which brings diversity into training. In Section~\ref{sec:comparison_state_of_art}, we show the impact of $p$ in terms of performance and speedup. 
 %To validate our strategy we leverage the largest Re-Identification dataset Veri-Wild~\cite{} which has $277,797$ images  in the training set from $30,000$ vehicles captured from $172$ different cameras in a $200km^{2}$ area in a city. Since MSMT17 is the largest person re-identification dataset, we also evaluate our strategy over it despite being almost ten times smaller than Veri-Wild. 
 
%We do not consider other large-scale datasets, such as ImageNet~\cite{deng2009imagenet}, %or Places~\cite{wang2020online} 
%since our goal is to tackle challenging re-identification scenarios with significant variations in light, points of view, resolution, occlusions, and background. Therefore, we employ large-scale Re-Identification datasets only. 

\subsection{Local Re-Ranking}
\label{subsec:local_reranking_clustering}

Prior art usually employs the $k$-Reciprocal Encoding algorithm~\cite{zhong2017re} (Full Re-Ranking -- FRR) to account for the context (neighborhood) of each sample. This helps to compute a more robust distance measure and allows hard positive samples to be closer in the feature space. Despite its effectiveness, we argue it might not be efficient, in terms of memory and space, when considering large-scale scenarios. 

To improve efficiency when dealing with large-scale datasets while keeping the advantages of neighborhood-based distance refinement, we propose a new Local Re-Ranking (LRR) algorithm. At a given step, it only considers the local neighborhood of two samples. The idea is to consider samples in common in given neighborhoods, avoiding the full comparison between all samples in the training set. Without loss of generality, consider a feature vector set $F_{m}$ ($1 \leq m \leq 3$) created after feature extraction by one of the backbones in Stage 2. We first calculate the $k$-Nearest Neighbors set $N(x_{i},k)$ for each sample $x_{i} \in F_{m}$, and the local distance matrix $D_{loc} \in R^{n \times k}$, where the $i$-th line is the distance between $x_{i}$ and each element in $N(x_{i},k)$ after applying the exponential decay function. That is, $D_{loc}(i,j) = e^{-d(x_{i}, x_{j})}$, where $x_{j} \in N(x_{i},k)$ and $d(.,.)$ is the Euclidean distance. The lower (or greater) the distance, the closer to one (or zero) they are after the exponential decay transformation. After that, we employ our neighborhood calculation based on the Jaccard distance to refine the distance between each $x_{i}$ and its nearest neighbors. For each sample $x_{j} \in N(x_{i}, k)$ we calculate the following sets:
\begin{equation}
\label{eq:intersection_set_equation}
    I(x_{i}, x_{j}) = \{p | p \in N(x_{i}, k) \wedge p \in N(x_{j}, k)\},
\end{equation}
\begin{equation}
\label{eq:just_in_Aset}
   E(x_{i}, x_{j}) = N(x_{i},k) \backslash I(x_{i},x_{j}).
\end{equation}
$I(x_{i}, x_{j})$ is the Inclusion set, which contains the common neighbor elements, and $E(x_{i})$ is the Exclusion set, which contains elements in $N(x_{i}, k)$ but not in $N(x_{j},k)$. Following~\cite{zhong2017re}, we assume the greater the cardinality of $I(x_{i}, x_{j})$, the more likely $x_{i}$ and $x_{j}$ are samples from the same class. In light of this, we propose the following to refine the distance:
\begin{equation}
\label{eq:sum_min}
   s_{min} = \sum_{p \in I(x_{i}, x_{j})} min(D_{loc}(i,p), D_{loc}(j,p)),
\end{equation}
\begin{equation}
\label{eq:sum_max}
   s_{max} = \sum_{p \in I(x_{i}, x_{j})} max(D_{loc}(i,p), D_{loc}(j,p)),
\end{equation}
\begin{equation}
\label{eq:sum_xi}
   s(x_{i}, x_{j}) = s_{i,j} = \sum_{p \in E(x_{i}, x_{j})} D_{loc}(i,p),
\end{equation}
\begin{equation}
\label{eq:sum_xj}
   s(x_{j}, x_{i}) = s_{j,i} = \sum_{p \in E(x_{j}, x_{i})} D_{loc}(j,p),
\end{equation}
\begin{equation}
\label{eq:iou_distance}
   D_{IoU}(x_{i}, x_{j}) = \frac{s_{min}}{s_{max} + s_{i,j} + s_{j,i}},
\end{equation}
\noindent where $s_{min}$ is the sum of the minimum values when comparing the distances of $x_{i}$ and $x_{j}$ to a common neighbor $p$ (the same for $s_{max}$ but considering maximum values), and $s_{i, j}$ (or $s_{j, i}$) is the sum of the distances between sample $x_i$ (or $x_j$) and its neighbors that are not in the intersection $I(x_i, x_j)$. 

In extreme cases, when $x_{i}$ and $x_{j}$ have all their neighbors in common, we have $I(x_{i}, x_{j}) = N(x_{i}, k) = N(x_{j}, k)$, $E(x_{i}, x_{j}) = E(x_{j}, x_{i}) = \emptyset$, $s_{i,j}= 0$ and $s_{j,i} = 0$, and Equation~\ref{eq:iou_distance} becomes $0 \leq D_{IoU}(x_{i}, x_{j}) = s_{min}/s_{max} \leq 1$. Conversely, when they do not have any neighbors in common, we have $I(x_{i}, x_{j}) = \emptyset$, $E(x_{i}, x_{j}) = N(x_{i}, k)$, $E(x_{j}, x_{i}) = N(x_{j}, k)$, $s_{min} = 0$, $s_{max} = 0$, $s_{i, j} \neq 0$ and $s_{j, i} \neq 0$, then $D_{IoU}(x_{i}, x_{j}) = 0$. Therefore, we see that $ 0 \leq D_{IoU}(x_{i}, x_{j}) \leq 1$, and the closer it is to 1, the more likely it is for $x_{i}$ and $x_{j}$ to be from the same class. Finally, to convert $D_{IoU}$ into a distance measure, we define the refined distance matrix $R$ as
\begin{equation}
\label{eq:relation_iou_distance}
  R(x_{i}, x_{j}) =
    \begin{cases}
      1 - D_{IoU}(x_{i}, x_{j}), & \text{if $x_{i} \in N(x_{j}, k)$}\\
      1, & \text{otherwise},
    \end{cases}       
\end{equation}
\noindent for each $x_{j} \in N(x_{i}, k)$. Note that $R \in \mathbb{R}^{n \times k}$ is used to perform clustering in further steps.

In terms of complexity, we compare LRR and FRR theoretically. %without considering hardware optimization. %, such as parallelism. 
In FRR, the authors first compute the $k$-Nearest Neighbors set for each training sample in $\mathcal{O}(n^{2}\log{n})$. Then, they calculate the $k$-reciprocal nearest neighbor set in $\mathcal{O}(nk^{2})$, and the incremental set in $\mathcal{O}(nk^{3}/2)$. The full distance matrix is created based on the incremental set in $\mathcal{O}(n^2)$, and the final Jaccard distances are computed in $\mathcal{O}(n^3/2)$. Finally, they perform the local query expansion in $\mathcal{O}(n^2k)$. Therefore, FRR's complexity is $\mathcal{O}(n^3 + n^{2}\log{n} + n^2k + n^2 + nk^{3} + nk^{2}) = \mathcal{O}(n^3)$. For LRR, we also calculate the $k$-Nearest Neighbors in $\mathcal{O}(n^{2}\log{n})$ and distance matrix $D_{loc}$ in $\mathcal{O}(nk)$. The intersection set $I(x_{i}, x_{j})$ and the sets $E(x_{i})$ and $E(x_{j})$ are computed in $\mathcal{O}(3k^2)$. Equations~\ref{eq:sum_min} and~\ref{eq:sum_max} can be calculated in a single pass in $\mathcal{O}(k)$. Equations~\ref{eq:sum_xi} and~\ref{eq:sum_xj} are computed in 
$\mathcal{O}(k)$, so the final complexity is $\mathcal{O}(3k^2 + 3k)$ for a single pair $x_{i}$ and $x_{j}$. Considering all possible pairs in $D_{loc}$ and the whole training set, the complexity is $\mathcal{O}(3nk^3 + 3nk^2)$. Therefore, LRR's total complexity is $\mathcal{O}(n^{2}\log{n} + nk^3 + nk^2 + nk) = \mathcal{O}(n^{2}\log{n})$ with $k \ll n$. 
%where the training set size $n$ is much greater than the nearest neighbor set size ($k << n$). 
Our model is more efficient and applicable to large-scale datasets, which is also corroborated by the practical time analysis (Section~\ref{sec:ablation_study}). %We also provide a Cython-based implementation for practical speedup.   

\subsection{Noise-Robust Density Scheduling}
Previous clustering-based methods often assume strong assumptions. For instance, when using DBSCAN, it is common to use some side information to define $\varepsilon$, such as holding it fixed or calculating it using extra information.
Some works select a fixed $\varepsilon$ parameter during the unsupervised training in the clustering step. As the backbone's weights are constantly changing, creating different feature spaces, a fixed clustering hyper-parameter is likely suboptimal. %In early iterations, the backbones are strongly biased by the camera view, i.e., feature vectors from images from the same identity and same camera tend to be grouped together. So, if the $\varepsilon$ is too high, clusters might include false-positive samples. 
In early iterations, the backbones are strongly biased by the camera view, i.e., feature vectors from images from the same identity and the same camera tend to be grouped together. Conversely, images from the same person from different cameras are farther away than images from different identities but with the same camera. So, if $\varepsilon$ is too high, clusters might include false-positive samples from the same camera.
As the training progresses, the backbones start to learn camera-invariant representations, and more cross-view images from the same class are included in the clusters, but $\varepsilon$ might not be large enough to include diverse examples from different cameras. Hence, methods that consider a fixed $\varepsilon$ face a noise/diversity trade-off.  

Other unsupervised methods set an optimal $\varepsilon$ per dataset~\cite{zhang2022implicit, peng2023adaptive, yin2023real} because each dataset has a different complexity level. %For instance, Market1501~\cite{zheng2015scalable} has 12,936 training images from 751 identities recorded from six different outdoor cameras on a university campus, while MSMT17~\cite{wei2018person} has 32,621 images from 1,041 identities recorded from fifteen cameras (twelve outdoors and three indoors) in an entrance place along different days and months. 
For instance, previous research has reported a lower $\varepsilon$ for the \texttt{Market1501} dataset, as it has less variability in terms of cameras and identities, and a greater $\varepsilon$ for \texttt{MSMT17} due to a higher camera diversity. However, this selection is unrealistic since the main assumption is the full absence of labels or any other side information. It is not trivial to find an optimal value when assuming no prior knowledge about the dataset complexity~\cite{hou2016dsets, sharma2017knn}. Aiming to propose a more general approach, closer to a real deployment scenario, we introduce an $\varepsilon$ scheduling scheme to address the diversity/noise trade-off during training, without requiring per-dataset tuning.

\begin{figure}[ht]
\centering
\includegraphics[width=1.7in]{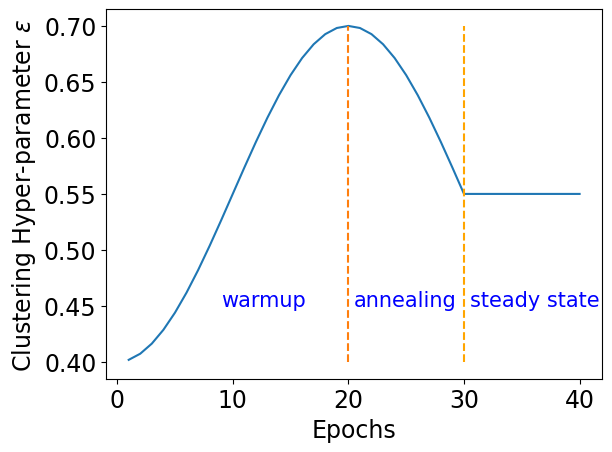}
\caption{Noise-Robust Density Scheduling during training for the $\varepsilon$ parameter. First, a \textit{warmup} is performed to address the diversity/noise trade-off. In the \textit{annealing} phase, we decrease $\varepsilon$ to reduce the influence of noisy data while keeping the diversity within the clusters. In the last phase, the \textit{steady state}, we keep the same $\varepsilon$ value until the end to encourage a stable behavior.}
\label{fig:eps_scheduling}
\end{figure}

We employ DBSCAN for clustering samples given their distances in matrix $R$ (Equation~\ref{eq:relation_iou_distance}). As distances are normalized between $0.0$ and $1.0$, the $\varepsilon$ value must be selected within this range. %A smaller value, as mentioned before, might create clusters with high true positive rates, but most of them will have images just from one camera since the background and camera view dominates the feature representation in early iterations. However, when the backbone is optimized to get samples from the same cluster closer and apart from the other clusters, it starts learning how to separate the samples from two (or more) different clusters that compress images from the same camera. Consequently, to some extent, the model alleviates the camera bias and starts to group more and more true positive cross-view images in the clusters. However, if $\varepsilon$ is fixed in a small value, it might not be enough to include in the same cluster most of the cross-view images since true positive cross-view images might be spread in the feature space. In counterpart, if the $\varepsilon$ has been large since the early iterations, it might include more cross-view images along the training, but it will also include more noisy samples, which can hinder the model's performance. 
We propose a novel $\varepsilon$ scheduling scheme, in which it starts from a low value and gradually increases, following a cosine scheduling, until half of the training epochs, as shown in Figure~\ref{fig:eps_scheduling}. We call this first phase the \textit{warmup}. As $\varepsilon$ is progressively increased, we allow the clustering algorithm to consider more cross-view images in a smoother manner. As previously reported by noise-robust feature learning methods~\cite{han2018co, yu2019does, song2022learning}, the method first learns from clean examples before starting to learn from the noisy ones. Since we start with high true-positive rate clusters in early iterations and smoothly increase the margin, the effect of noisy samples is alleviated even if the noise ratio starts to increase.

If we keep $\varepsilon$ in its highest value until the end of the training, the noisy samples are overemphasized, and the backbone overfits~\cite{han2018co, yu2019does, han2020survey, song2022learning}. To tackle this, we gradually decrease $\varepsilon$ to avoid adding too many noisy samples to the clusters. We call this the \textit{annealing} phase. After the model reaches $75\%$ of training epochs, $\varepsilon$ is kept in a plateau until the end of the training, in a \textit{steady state} phase. As the backbone learns to group the cross-view images in the first half of the training, the cross-view image representations tend to be closer at this stage. Thus, the backbone can keep learning from diverse samples as $\varepsilon$ is gradually decreased, alleviating the impact of the noisy samples. While our experiments use a fixed number of epochs, the scheduling would also allow a temporal stopping criterion. %, inspired by ~\cite{han2018co}. They propose supervised training where they select small-loss samples from one peer network to train the other. 
%They gradually decrease the small-loss threshold to avoid the influence of noisy samples and get more and more confident samples for training. 
%In counterpart, we redesigned the solution for the fully-unlabeled scenarios by not working in the loss value, but in the neighborhood density definition to obtain a model operating in fully-unlabeled complex scenarios, avoiding the hard and challenging task of selecting a proper $\varepsilon$ for each application. It is important to note that \textbf{our whole formulation does not consider the camera labeling}. It learns to group cross-view images in a fully-unsupervised and automatic way.  

\subsection{Co-training}
\label{sec:co_training}
After defining $\varepsilon$ for the current epoch and performing clustering for each of the $M$ backbones, we have a sequence $Y$ of $M$ pseudo-label sets. It is defined as $Y = (Y_{m})_{m=1}^{M}$, where $Y_{m}$ is the set of pseudo-labels generated by clustering the features extracted by the $m$-th backbone. 

In previous Person Re-Identification works~\cite{ge2020mutual, zhai2020multiple, zhai2022population}, co-training is considered but with a mechanism involving hard and soft supervision among the backbones, often leading to a complex loss function and optimization process. In our case, we propose a parameter-free co-training strategy by permuting the generated pseudo-labels among the backbones. Formally, we generate a random permutation $Y^{'}$ of the sequence $Y$, giving that $Y^{'}_{m} \neq Y_{m}$, i.e., a backbone should never be supervised by its own pseudo-label set.

After that, each backbone $f_{m}$ is trained with a permuted pseudo-label set $Y^{'}_{m}$ (Stage 6, Figure~\ref{fig:full_pipeline}). As an illustrative example, let $Y_{1} = (-1,0,1)$ be the pseudo-labels generated by $f_1$ and $Y_{2} = (0, 0, 1)$ by $f_2$. If they are permuted, then $Y^{'}_{1}= (0, 0, 1)$ and $Y^{'}_{2}= (-1,0,1)$ are the sets carried on to the next stage for $f_1$ and $f_2$, respectively.
%is given in Table~\ref{tab:co_training_example}. 
This encourages complementary knowledge sharing among the backbones and alleviates error amplification since a backbone is supervised by one of other $M-1$ pseudo-labels sets. This solution outperforms all co-training techniques used in PReID methods, in the most complex scenarios. The permuted labeling is employed in the next stage to optimize the loss functions.

\subsection{Optimization and self-ensembling}
The loss function (Stage 7, Figure~\ref{fig:full_pipeline}) comprises two terms: the proxy loss $L_{proxy}$ and the hard loss $L_{hard}$. For one backbone $f_{m}$, we first randomly select one sample per cluster in $Y^{'}_{m}$ to be the proxy of the cluster. As $Y^{'}_{m}$ contains samples from $c_{m}$ different clusters, we have the set $P_{m} = \{p_{m}^{1}, ..., p_{m}^{c_{m}} \}$ with $c_{m}$ proxies. Given a batch $B$ of norm-1 image feature vectors, we calculate the proxy loss as follows:
\begin{equation}
\label{eq:proxy_loss}
   L_{proxy} = -\frac{1}{|B|}\sum_{v \in B}log \frac{exp(v\cdot p_{m}^{+}/\tau)}{\sum_{q \in P_{m}} exp(v\cdot q/\tau)},
\end{equation}
\noindent where $p_{m}^{+}$ is the proxy of the cluster of feature vector $v$, $v\cdot q$ is the dot product between $v$ and $q$, and $\tau$ is a temperature hyper-parameter to control the shape of the distribution. %All vectors have norm 1.

Moreover, we consider the hard-positive and hard-negative samples in the batch~\cite{hermans2017defense} to increase feature representation:
\begin{equation}
\label{eq:batch_hard}
   L_{hard} = -\frac{1}{|B|}\sum_{v \in B}log \frac{exp(v\cdot v_{pos}/\tau)}{exp(v\cdot v_{pos}/\tau) + exp(v \cdot v_{neg}/\tau)},
\end{equation}
\noindent where $v_{pos}$ is the furthest sample in $v$'s cluster and in $B$, and $v_{neg}$ is the closest sample to $v$ in $B$ from a different cluster. 

$L_{proxy}$ is a more global loss because it considers all proxies in the calculation, while $L_{hard}$ enforces a more local view by considering just in-batch samples. The final loss function is
\begin{equation}
\label{eq:final_loss}
   L_{final} = L_{proxy} + \lambda L_{hard},
\end{equation}
\noindent where $\lambda$ controls the contributions of $L_{hard}$. In the Supplementary Material, we present the impact of $\tau$ and $\lambda$. 

To avoid noise amplification during training, we also perform the self-ensembling of each model's weights as usually done in previous methods~\cite{he2020momentum, caron2021emerging}. For each backbone $f_m$, we keep a self-ensembled model with parameters $\Theta_{m}^{t}$. They are updated as $\Theta_{m}^{t}:= \beta\Theta_{m}^{t-1} + (1-\beta)\theta_{m}^{t}$, where $\theta_{m}^{t}$ are the parameters of backbone $f_m$ optimized by Equation~\ref{eq:final_loss}, and $\beta$ is an inertia hyper-parameter set to $0.999$ as in~\cite{he2020momentum}. % in all models and experiments.

\subsection{Inference}

After the training pipeline, the inference is made by ranking all gallery samples based on the distance to a query sample. We extract feature vectors for all \textit{query} and \textit{gallery} sets using the self-ensembled models. These sets are denoted by $F_{q}^{m}$ and $F_{g}^{m}$, respectively, with $1 \leq m \leq M$. We calculate pairwise distances between samples of $F_{q}^{m}$ and $F_{g}^{m}$, resulting in a distance matrix $D_{q2g}^{m} \in \mathbb{R}^{|F_{q}^{m}| \times |F_{g}^{m}|}$. %where $|Q|$ and $|G|$ are the number of samples in the query and gallery sets. 

A final distance matrix $\overline{\rm D}_{q2g}$ is obtained by averaging all matrices element-wise:
\begin{equation}
\label{eq:distance_ensemble_evaluation}
    \overline{\rm D}_{q2g} = \frac{1}{M}\sum_{m=1}^{M}D_{q2g}^{m}.
\end{equation}
Each row of $\overline{\rm D}_{q2g}$ holds the distances from a query to the gallery samples. We sort these distances to get the closest class to the query sample. As done in previous works, we remove gallery images with the same class and camera of the query to assess performance in a true cross-camera scenario.

\section{Experiments and Results}
\label{sec:experiments}

In this section, we present the datasets, metrics, and results of our proposed pipeline compared to the prior art in the fully-unsupervised Person and Vehicle ReID problem.

\subsection{Datasets and Implementation Details}
\label{subsec:datasets}
We evaluate our method in two Person ReID, \texttt{Market1501}~\cite{zheng2015scalable} and \texttt{MSMT17}~\cite{wei2018person}, and three Vehicle ReID datasets, \texttt{Veri}~\cite{liu2016large}, \texttt{VehicleID}~\cite{liu2016deep}, and \texttt{Veri-Wild}~\cite{lou2019veri}, as summarized in Table~\ref{tab:dataset_information}. The last two are large-scale datasets with more than $100K$ images in the training set and with three evaluation scenarios. \texttt{Veri-Wild}~\cite{lou2019veri} is the most challenging one compressing 174 cameras while \texttt{Market1501}, \texttt{MSMT17}, and \texttt{Veri} compress six, fifiteen and twenty cameras respectively. The number of cameras in \texttt{VehicleID} is not informed.

\begin{table}[ht]
\caption{Information about the employed datasets.}
\centering
\begin{tabular}{P{1.0cm}|P{0.7cm}|P{0.8cm}|P{0.7cm}|P{0.8cm}|P{0.7cm}|P{0.8cm}}
&
\multicolumn{2}{|c|}{\textbf{train}} &
%\multicolumn{2}{|c}{\textbf{MSMT17}} &
\multicolumn{2}{|c}{\textbf{gallery}} &
\multicolumn{2}{|c}{\textbf{query}} \\ \hline
& \#IDs & \#Images & \#IDs & \#Images & \#IDs & \#Images \\ \hline
Market & 751 & 12,936 & 751 & 15,913 & 750 & 3,368\\
MSMT17 & 1,041 & 32,621 & 3,060 & 82,161 & 3,060 & 11,659\\
Veri & 576 & 37,778 & 200 & 11,579 & 200 & 1,678\\ \hline
\multirow{3}{*}{VehicleID} & \multirow{3}{*}{13,164} & \multirow{3}{*}{113,346} & 800 & 800 & 800 & 5,693 \\
& & & 1,600 & 1,600 & 1,600 & 11,777\\
& & & 2,400 & 2,400 & 2,400 & 17,377\\
\hline
\multirow{3}{*}{Veri-Wild} & \multirow{3}{*}{30,671} & \multirow{3}{*}{277,797} & 3,000 & 38,861 & 3,000 & 3,000 \\
& & & 5,000 & 64,389 & 5,000 & 5,000\\
& & & 10,000 & 128,517 & 10,000 & 10,000\\
\hline
\end{tabular}
\label{tab:dataset_information}
\end{table}

For evaluation, we calculate the Cumulative Matching Curve (CMC), from which we report Rank-1 (R1), Rank-5 (R5), Rank-10 (R10), and mean Average Precision (mAP).

%\subsection{Implementation Details}

We adopt ResNet50~\cite{he2016deep}, DenseNet121~\cite{huang2017densely}, and OSNet~\cite{zhou2019omni} pre-trained in ImageNet~\cite{deng2009imagenet} as our backbones. We use Pytorch~\cite{NEURIPS2019_9015}, Torchreid~\cite{torchreid}, and FAISS~\cite{johnson2019billion} as supporting libraries. %as commonly done by the prior art. 
We first pre-train all backbones in each Person/Vehicle ReID dataset using the self-supervised Barlow Twins strategy. We randomly select one backbone to extract features and obtain the top-$p\%$ ($p$ can be $25\%$, $50\%$, $75\%$ or $100\%$) samples in LNS, every three epochs. We linearly warm up the learning rate from $3.5e-5$ to $3.5e-4$ in the first 10 epochs and keep it fixed for the remaining training in a total of $40$ epochs. We use the Adam~\cite{kingma2014adam} optimizer with weight decay set to $5e-4$. The batches are created by randomly sampling $16$ pseudo-identities (clusters) and $12$ images from each. We sample batches until all pseudo-identities are covered, then we repeat this process five times before going to the next epoch. Before each epoch, we renew the class proxies. The parameter $\tau$ is set to $0.04$ and $\lambda$ to $0.5$ in all experiments. 

We trained the models in five TITAN RTX GPUs with 24GB of memory each. One GPU is left just for Local Re-Ranking and clustering, and the others are used for training. Further implementation details are in the Supplementary Material.

\subsection{Comparison to State-of-The-Art methods}
\label{sec:comparison_state_of_art}

% We compare our method with the prior art, and results are shown in Tables~\ref{tab:state_of_art_reid_fully_unsupervised},~\ref{tab:state_of_art_reid_ensembles},~\ref{tab:state_of_art_vehicle},~\ref{tab:vehicleID}, and~\ref{tab:veri_wild}. 

Table~\ref{tab:state_of_art_reid_fully_unsupervised} shows our method compared to prior fully-unsupervised Person ReID models that, \textbf{like ours, do not consider camera labels}. Results for other methods that use side information are in the Supplementary Material. We flag the methods that tune the clustering parameter per dataset because they cannot be used in a realistic fully-unsupervised scenario (column \textit{CPD}). All comparison tables show mAP and R1. R5 and R10 are shown in the Supplementary Material.  

AdaMG~\cite{peng2023adaptive} achieves good results in \texttt{Market} and \texttt{MSMT17} by setting $\varepsilon = 0.5$ and $\varepsilon = 0.7$, respectively, but with a fixed $\varepsilon$, they face a huge performance drop~\cite{peng2023adaptive}. With our $\varepsilon$ scheduling scheme, we outperformed AdaMG by $1.2$ and $0.1$ percentage points (p.p.) in mAP and R1, respectively, in \texttt{Market}, and by $5.2$ and $4.6$ p.p. in the challenging \texttt{MSMT17}.

% In Table~\ref{tab:state_of_art_reid_fully_unsupervised}, we compared our models to prior Person Re-Identification datasets.  We compare ours to fully-unsupervised models \textbf{without considering camera labeling}. Further models considering camera labeling are compared in the Supplementary Material. It is important to point out some of the methods in Table~\ref{tab:state_of_art_reid_fully_unsupervised} tune the $\varepsilon$ parameter of DBSCAN (or the clustering hyper-parameter involved in the pipeline) in each dataset to achieve the best performance, which is unrealistic in a fully-unsupervised scenario since it does not have any kind of validation data to properly pick the best clustering hyper-parameters. For instance, the best model in prior art AdaMG~\cite{peng2023adaptive} achieves the performance in Market and MSMT17 by setting $\varepsilon = 0.5$ and $\varepsilon = 0.7$. Besides, taking their own ablation study table (Table IX in~\cite{peng2023adaptive}) it is possible to see that if they keep $\varepsilon = 0.5$ for both datasets, they face a huge performance dropping in mAP ($38.0\%$ to $11.4\%$) and in R1 ($66.3\%$ to $32.8\%$) in MSMT17 which is the most challenging Person ReID dataset. Our model in counterpart does not rely on specific tuning in any dataset, and the same $\varepsilon$ scheduling is employed in all datasets. We outperformed AdaMG by $1.2$ and $0.1$ in mAP and R1, respectively, in Market, and by $5.2$ and $4.6$ in mAP and R1 in the challenging MSMT17. 

Due to the proposed Local Re-Ranking, our model has a lower Re-Ranking memory footprint ($\mathcal{O}(kN)$ with $k \ll N$) compared to the best methods HHCL, GRACL, and AdaMG ($\mathcal{O}(N^2)$) and still outperforms them in mAP and R1 in \texttt{Market} and in all metrics in \texttt{MSMT17}. Our method also outperforms all other methods~\cite{nikhal2021unsupervised, zheng2022clustering, ding2019dispersion, sun2021unsupervised, yin2021multi, pan2020unsupervised, wang2020unsupervised, nikhal2022multi, yu2022graph, tang2021fully, zeng2020hierarchical, tang2021unsupervised, tang2022unsupervised, zhao2022unsupervised, yin2022unsupervised, si2023diversity, sun2022hybrid, li2023multi, pang2023reliability, zhang2022unsupervised, jin2022meta, zhang2022graph, wang2023relation} by a large margin. 

% Our method also ranks second place in MSMT17 using just $75\%$ of the data in each iteration, while the remaining methods start with $100\%$, showing our method is able to mine useful patterns even with a reduced amount of data. Further analysis is given in Section~\ref{sec:result_speedup_local_sampling}. 

Overall, our method provides the best mAP and R1 without relying upon any dataset-specific hyper-parameter and with a more efficient Re-Ranking strategy. 

\begin{table}[ht]
%\vspace{-55pt}
\caption{Comparison with relevant fully-unsupervised Person ReID methods. The best result is highlighted in \bv{blue}, the second best in \gv{green}, and the third in \ov{orange}. RRMC means Re-Ranking Memory Complexity and CPD (Cluster Parameter per Dataset) indicates if the method relies on specific clustering parameters per dataset. (p\%) means that p\% of all data points are sampled in LNS at each epoch.}
\label{tab:state_of_art_reid_fully_unsupervised}
\centering
\begin{tabular}{p{1.6cm}| p{1.2cm}|P{0.75cm}|P{0.4cm}|p{0.4cm}|p{0.4cm}|p{0.4cm}|p{0.4cm}}
\hline
\multicolumn{1}{c|}{} &
\multicolumn{1}{|c|}{} &
\multicolumn{1}{|c|}{} &
\multicolumn{1}{|c|}{} &
\multicolumn{2}{|c|}{\textbf{Market}} &
\multicolumn{2}{|c}{\textbf{MSMT17}} \\ \hline
%\multicolumn{14}{|c|}{\textit{Fully Unsupervised}}  \\ \hline
Method & Reference & RRMC & CPD & mAP & R1
& mAP & R1 \\ \hline
ABMT~\cite{chen2020enhancing} & WACV'20 & $\mathcal{O}(N^{2})$ & \gv{No} & 65.1 & 82.6 & - & -\\
SpCL~\cite{ge2020self} & NeurIPS'20 & $\mathcal{O}(N^{2})$ & \gv{No} & 73.1 & 88.1 & 19.1 & 42.3\\
GCL+~\cite{chen2022learning} & TPAMI'22 & $\mathcal{O}(N^{2})$ & \gv{No} & 69.3 & 89.0 & 22.0 & 47.9 \\ 
GSam~\cite{han2022rethinking} & TIP'22 & $\mathcal{O}(N^{2})$ & \gv{No} & 79.2 & 92.3 & 24.6 & 56.2 \\
%SpCL-IBN~\cite{ge2020self} & NeurIPS'20 & 73.8 & 88.4 & 95.3 & 97.3 & - & - & - & - & 24.0 & 48.9 & 61.8 & 67.1 \\
RLCC~\cite{zhang2021refining} & CVPR'21 & $\mathcal{O}(N^{2})$ & \gv{No}&  77.7 & 90.8 & 27.9 & 56.5 \\
%MaskPre~\cite{yin2022unsupervised} & PR'22 & - & \gv{No} & 77.7 & 90.4 & 96.4 & 97.9 & 26.7 & 58.5 & 69.3 & 73.8 \\
HCLP~\cite{zheng2021online} & ICCV'21 & - & \gv{No} & 78.1 & 91.1 & 26.9 & 53.7\\
%DFC~\cite{si2023diversity} & IPM'23 & - & \gv{No} &  78.6 & 90.8 & 97.5 & - & - & - & - & - \\
ICE~\cite{Chen_2021_ICCV} & ICCV'21 & $\mathcal{O}(N^{2})$ & \gv{No} &79.5 & 92.0 & 29.8 & 59.0\\
%MGCE-HCL~\cite{sun2022hybrid} & ACPR'22 & - & \gv{No} & 79.6 & 92.1 & - & - & - & - & - & - \\
%MPC~\cite{li2023multi} & CVIU'23 &$\mathcal{O}(|C_{i}|^{2})$ & \gv{No} & 77.4 & 90.9 & 96.4 & 97.6 & - & - & - & - \\
CACL~\cite{li2022cluster} & TIP'22 & - & \gv{No} & 80.9 & 92.7 & 23.0 & 48.9\\
HDCRL~\cite{cheng2022hybrid} & TIP'22 & - & \gv{No} & 81.7 & 92.4 & 24.6 & 50.2\\
PPLR~\cite{cho2022part} & CVPR'22 & $\mathcal{O}(N^{2})$ & \gv{No} & 81.5 & 92.8 & 31.4 & 61.1\\
%RMCL~\cite{pang2023reliability} & KBS'23 & $\mathcal{O}(N^{2})$ & \gv{No} & 81.7 & 93.0 & 97.6 & 98.4 & 32.5 & 62.3 & 73.6 & 78.0 \\
%Zhang et. al.~\cite{zhang2022unsupervised} & AI'22 & $\mathcal{O}(N^{2})$ & \gv{No} & 83.0 & 93.2 & 97.3 & - & - & - & - & - \\
RTMem~\cite{yin2023real} & TIP'23 & - & \rv{Yes} & 83.0 & 92.8 & 32.8 & 57.1\\
CCons~\cite{dai2022cluster} & ACCV'22 & $\mathcal{O}(N^{2})$ & \gv{No} & 83.0 & 92.9 & 33.0 & 62.0 \\
%MCL~\cite{jin2022meta} & ICM'22 & $\mathcal{O}(N^{2})$ & \rv{Yes} & 83.3 & 93.0 & - & - & 33.4 & 62.9 & - & - \\
%GCM~\cite{zhang2022graph} & SIVP'22 & $\mathcal{O}(N^{2})$ & \gv{No} & 83.4 & 93.3 & - & - & - & - & - & - \\
ISE~\cite{zhang2022implicit} & CVPR'22 & - & \rv{Yes} & \gv{84.7} & \bv{94.0} & 35.0 & 64.7\\
%Zheng~\cite{zheng2022multi} & ArXiv'22 & $\mathcal{O}(N^{2})$ & \gv{No} & 83.1 & 92.8 & 97.1 & 98.0 & 35.6 & 63.8 & 75.3 & 79.5 \\
HHCL~\cite{hu2021hard} & NIDC'21 & $\mathcal{O}(N^{2})$ & \gv{No} & 84.2 & \ov{93.4} & - & -\\
GRACL~\cite{zhang2022global} & TCSVT'22 & $\mathcal{O}(N^{2})$ & \gv{No} &  83.7 & 93.2 & 34.6 & 64.0\\
%CGCL~\cite{miao2022confidence} & ArXiv'22 & - & \gv{No} & 85.3 & 94.2 & 97.6 & 98.5 & 34.6 & 63.4 & 74.6 & 79.3 \\
%NCPLR~\cite{cheng2022neighbour} & ArXiv'22 & - & \rv{Yes} & 86.3 & 94.3 & 98.0 & 98.7 & 35.7 & 66.3 & 76.9 & 80.6 \\
AdaMG~\cite{peng2023adaptive} & TCSVT'23 &  $\mathcal{O}(N^{2})$ & \rv{Yes} & \ov{84.6} & \gv{93.9} & \ov{38.0} & \ov{66.3}\\
%DCCT~\cite{chen2022dual} & ArXiv'22 &  $\mathcal{O}(N^{2})$ & \rv{Yes} & 86.3 & 94.4 & 97.7 & 98.5 & 41.8 & 68.7 & 79.0 & 82.6 \\ 
\hline
%\textbf{Ours} & & 84.4 & 93.2 & 96.7 & 97.6 & 45.3 & 71.9 & 81.0 & 83.9\\
%\textbf{Ours (best $k$)} & & 85.7 & 93.9 & 97.2 & 98.0 &  &  &  & \\
\textbf{Ours (25\%)} &  & $\mathcal{O}(kN)$& \gv{No} & - & - & 32.0 & 60.5\\
\textbf{Ours (50\%)} & & $\mathcal{O}(kN)$ & \gv{No} & - & - & 24.3 & 50.4\\
\textbf{Ours (75\%)} & & $\mathcal{O}(kN)$ & \gv{No} & 82.9 & 92.6 & \gv{39.3} & \gv{67.3}\\
\textbf{Ours (100\%)} &  & $\mathcal{O}(kN)$ & \gv{No} & \bv{85.8} & \bv{94.0} & \bv{43.2} & \bv{70.9}\\

\hline
\end{tabular}
\end{table}

\begin{table}[ht]
%\vspace{-55pt}
\caption{Comparison with relevant ensemble-based Person ReID methods. ``\# BB'' shows the number of backbones used in training or evaluation, ``Src'' means if a labeled dataset has been used to initialize the model before training: ``M'' for \texttt{Market1501} and ``D'' for \texttt{DukeMTMC-ReID}. The best result is highlighted in \bv{blue} and the second best in \gv{green}.}.
\label{tab:state_of_art_reid_ensembles}
\centering
\begin{tabular}{p{1.2cm}|P{0.95cm}|P{0.7cm}|P{0.25cm}|P{0.4cm}|P{0.4cm}|P{0.25cm}|P{0.4cm}|P{0.3cm}}
\hline
\multicolumn{3}{c|}{} &
\multicolumn{3}{|c|}{\textbf{Market}} &
%\multicolumn{1}{|c|}{} &
\multicolumn{3}{|c}{\textbf{MSMT17}} \\ \hline
Method & Reference & \# BB & Src & mAP & R1
& Src & mAP & R1 \\ \hline
ACT~\cite{yang2020asymmetric} & AAAI'20 & 2 & D & 60.6 & 80.5 & - & - & - \\
MMT~\cite{ge2020mutual} & ICLR'20 & 2 & D & 71.2 & 87.7 & M & 22.9 & 49.2\\ 
MEB~\cite{zhai2020multiple} & ECCV'20 & 3 & D & 76.0 & 89.9 & - & - & -\\
UST~\cite{9521886} & TIFS'21 & 3 & D & 78.4 & 92.9 & M & 33.2 & 62.3\\
ESSL~\cite{bertocco2022reasoning} & TIFS'23 & 3 & - & 83.4 & 92.9 & - & 42.6 & 68.2 \\ 
PEG~\cite{zhai2022population} & IJCV'22 & 8 & - & \bv{87.1} & \bv{94.6} & - & \gv{41.8} & \gv{69.1}\\
\hline
\textbf{Ours} & & 3 & - & \gv{85.8} & \gv{94.0} & - & \bv{43.2} & \bv{70.9}\\
\hline
\end{tabular}
\end{table}

Since our method is based on three different architectures, we also compare it to prior ensemble-based methods (Table~\ref{tab:state_of_art_reid_ensembles}). We obtain the second-best performance in \texttt{Market}. However, the best method, PEG~\cite{zhai2022population}, utilizes 8 backbones, which include ResNet50 and DenseNet121, like ours. Our third backbone, OSNet, has a lower memory footprint compared to the other backbones. They also employ a complex evolutionary-based strategy and co-training where the backbones are selected at different moments with different losses supervising each other. Our results were achieved with just three backbones and a simpler and parameter-free co-training. In \texttt{MSMT17}, which is more complex than \texttt{Market}, we outperform PEG by $1.4$ and $1.8$ p.p. in mAP and R1, respectively. Therefore, with a simpler strategy, we can still take advantage of the complementary knowledge from different backbones and outperform prior art in more complex scenarios.    

In the \texttt{Veri} dataset (Table~\ref{tab:state_of_art_vehicle}), we have the best R1 among all methods. Although we obtain the second-best result considering mAP, MSCL has a much lower R1 score, and they present a higher memory footprint for Re-Ranking. We are also a margin better than BUC~\cite{lin2019bottom}, MMCL~\cite{wang2020unsupervised}, and SSML~\cite{yu2021unsupervised}.

In the large-scale and challenging \texttt{VehicleID} (Table~\ref{tab:vehicleID}), we scored first in R1 in the most difficult scenario (TS = 2400) along with CCons~\cite{dai2022cluster}. However, CCons leverages a more memory-complex Re-Ranking. %, and lower by just by $0.2$ in mAP. 
In the other scenarios, we scored first or, at least, second place. %However, CCons leverages a more memory-complex Re-Ranking, and we overcome them in all other datasets by a large margin.
The results for \texttt{Test Size = 800} are presented in the Supplementary Material.

The \texttt{Veri-Wild} dataset has been less employed in the fully-unsupervised scenario. Usually, prior art utilizes camera labels, as shown in the Supplementary Material. Considering the few methods that, like ours, do not consider camera labels, we outperform them in the three evaluation scenarios in all metrics (Table~\ref{tab:veri_wild}). More specifically, in the most challenging setup (\texttt{VW-Large}), we outperform prior art by $4.5$ and $1.7$ p.p., in mAP and R1, respectively, with $75\%$ of the data. Furthermore, we provide the second- and third-best results with $100\%$ and $50\%$ of the data, respectively. The results for \texttt{Veri-Wild-Small} are presented in the Supplementary.

%Since the camera labeling has a strong capacity to help the models even without identity annotation, just a few models have reported performance in the fully-unsupervised scenario, as seen in Table~\ref{tab:veri_wild}, which is the focus of our work. Our method outperforms prior art in the three evaluation scenarios in all metrics. More specifically, in the most challenging setup (``Veri-Wild (Large)''), we outperform prior art by $4.2$ and $1.5$ in mAP and R1, respectively, with $100\%$ of data, and we can get even further improvement with $75\%$ of the data. 

\begin{table}[ht]
\caption{Comparison with relevant fully-unsupervised Vehicle ReID methods in \texttt{Veri776}. The best result is highlighted in \bv{blue}, the second best in \gv{green}, and the third in \ov{orange}. RRMC means Re-Ranking Memory Complexity, and CPD (Cluster Parameter per Dataset) indicates if the method relies on specific clustering parameters per dataset. (p\%) means that p\% of all data points are sampled in LNS at each epoch.}
\label{tab:state_of_art_vehicle}
\centering
\begin{tabular}{p{1.8cm}|p{1.2cm}|P{0.75cm}|P{0.4cm}|p{0.45cm}|p{0.40cm}|p{0.40cm}}
\hline
\multicolumn{1}{c|}{} &
\multicolumn{1}{|c|}{} &
\multicolumn{1}{|c|}{} &
\multicolumn{1}{|c|}{} &
\multicolumn{3}{|c}{Veri} \\
\hline
%\multicolumn{1}{|c|}{} &
%\multicolumn{1}{|c|}{} &
%\multicolumn{1}{|c|}{} &
%\multicolumn{1}{|c|}{} &
%\multicolumn{3}{|c|}{\textit{Part-based Models}} \\ \hline
%Method & Reference & RR-MC & CPPD? & mAP & R1 & R5  \\ \hline
%PPLR~\cite{cho2022part} & CVPR'22 & $\mathcal{O}(N^{2})$ & \gv{No} & 41.6 & 85.6 & 91.1\\ 
%MLPL~\cite{he2022multi} & TVT'22 & - & \gv{No} & 45.1 & 88.3 & 91.1\\ \hline
%\multicolumn{1}{|c|}{} &
%\multicolumn{1}{|c|}{} &
%\multicolumn{1}{|c|}{} &
%\multicolumn{1}{|c|}{} &
%\multicolumn{3}{|c|}{\textit{Attribute-based Models}} \\ \hline
%Method & Reference & RR-MC & CPPD? & mAP & R1 & R5\\ \hline
%VRPRD~\cite{bashir2019vr} & PR'19 & - & \gv{No} & 40.1 & 83.2 & 91.1 \\ \hline
%\multicolumn{1}{|c|}{} &
%\multicolumn{1}{|c|}{} &
%\multicolumn{1}{|c|}{} &
%\multicolumn{1}{|c|}{} &
%\multicolumn{3}{|c|}{\textit{Single-part-based Models}} \\ \hline
Method & Reference & RRMC & CPD & mAP & R1 & R5  \\ \hline
%BUC~\cite{lin2019bottom} & AAAI'19 & - & \gv{No} & 21.2 & 54.7 & 70.4  \\
%PAL~\cite{peng2020unsupervised} & ArXiv'20 & - & \gv{No} & 42.0 & 68.2 & 79.9 \\
%MMCL~\cite{wang2020unsupervised} & CVPR'20 & - & \gv{No} & 24.2 & 71.8 & 75.9 \\
%SSML~\cite{yu2021unsupervised} & IROS'21 & - & \gv{No} & 26.7 & 74.5 & 80.3\\
SpCL~\cite{ge2020self} & NeurIPS'20 & $\mathcal{O}(N^{2})$ & \gv{No} & 36.9 & 79.9 & 86.8 \\
GRACL~\cite{zhang2022global} & TCSVT'22 & $\mathcal{O}(N^{2})$ & \gv{No} & 39.4 & 82.9 & - \\
RLCC~\cite{zhang2021refining} & CVPR'21 & $\mathcal{O}(N^{2})$ & \gv{No} & 39.6 & \ov{83.4} & 88.8\\
CCons~\cite{dai2022cluster} & ACCV'22 & $\mathcal{O}(N^{2})$ & \gv{No} & 40.8 & \gv{86.2} & \bv{90.5}\\
AdaMG~\cite{peng2023adaptive} & TCSVT'23 & $\mathcal{O}(N^{2})$ & \rv{Yes} &  41.0 & \gv{86.2} & \gv{90.6} \\ 
RTMem~\cite{yin2023real} & TIP'23 & - & \rv{Yes} & \gv{41.8} & 81.6 & 87.0\\
%NCPLR~\cite{cheng2022neighbour} & ArXiv'22 &  - & \rv{Yes} & 42.0 & 85.8 & 90.5\\
MSCL~\cite{wang2022unsupervised} & SVIP'22 & $\mathcal{O}(N^{2})$ & \gv{No} &\bv{45.9} & 81.2 & - \\ 
\hline
\textbf{Ours (25\%)} & & $\mathcal{O}(kN)$ & \gv{No} & 28.0 & 66.8 & 72.2  \\
\textbf{Ours (50\%)} & & $\mathcal{O}(kN)$ & \gv{No} & 40.8 & 84.5 & 88.6  \\
\textbf{Ours (75\%)} & & $\mathcal{O}(kN)$ & \gv{No} & \ov{41.7} & \gv{86.2} & \ov{89.9}  \\
\textbf{Ours (100\%)} & & $\mathcal{O}(kN)$ & \gv{No} & 41.3 & \bv{86.3} & \ov{89.9} \\
\hline
\end{tabular}
\end{table}

\begin{table}[ht]
\caption{Comparison with relevant fully-unsupervised Vehicle ReID methods in \texttt{VehicleID}. The best result is highlighted in \bv{blue}, the second best in \gv{green}, and the third in \ov{orange}. RRMC means Re-Ranking Memory Complexity and CPD (Cluster Parameter per Dataset) indicates if the method relies on specific clustering parameters per dataset. Methods with * are reproduced from~\cite{he2022multi}. Speedup values are measured in comparison to the version with $100\%$ of the data. (p\%) means that p\% of all data points are sampled in LNS at each epoch. TS holds for Test Size.}
\label{tab:vehicleID}
\centering
\begin{tabular}{p{1.58cm}| P{1.0cm}|P{0.75cm}|P{0.40cm}|p{0.45cm}|p{0.45cm}|p{0.45cm}|p{0.45cm}}
\hline
\multicolumn{1}{c|}{} &
\multicolumn{1}{|c|}{} &
\multicolumn{1}{|c|}{} &
\multicolumn{1}{|c|}{} &
%\multicolumn{3}{|c|}{\textbf{Test size  = 800}} &
\multicolumn{2}{|c|}{\textbf{TS = 1600}} &
\multicolumn{2}{|c}{\textbf{TS = 2400}} \\
\hline
%\multicolumn{14}{|c|}{\textit{Camera-based}} \\ \hline
%\multicolumn{14}{|c|}{\textit{Segmentation-based}} \\ \hline
%\multicolumn{13}{|c|}{\textit{Part-based Models}} \\ \hline
%Method & Reference & RR-MC & CPD & mAP & R1 & R5 & mAP & R1 & %R5 & mAP & R1 & R5 \\ \hline
%MLPL~\cite{he2022multi}* & TVT'22 & - & \gv{No} & 65.3 & 61.1 & %69.8 & 62.7 & 57.3 & 68.4 & 59.6 & 52.4 & 66.2 \\ \hline
%\multicolumn{13}{|c|}{\textit{Fully Unsupervised}} \\ \hline
Method & Reference & RRMC & CPD & mAP & R1 & mAP & R1\\ \hline
BUC~\cite{lin2019bottom}* & AAAI'19 & - & \gv{No} & 46.2 & 45.9 & 42.4 & 39.7\\
%PAL~\cite{peng2020unsupervised}* & ArXiv'20 & - & \gv{No} & 53.5 & 50.3 & 64.9 & 48.1 & 44.3 & 60.9 & 45.1 & 41.1 & 59.1 \\ 
MAC~\cite{zhu2022manifold} & KBS'22 & - & \gv{No} & 51.9 & 47.5 & 47.4 & 44.4\\
%MSCL~\cite{wang2022unsupervised} & SVIP'22 & $\mathcal{O}(N^{2})$ & \gv{No} &  - & - & - & - & - & - & 52.9 & \bv{51.8} & - \\
SpCL~\cite{ge2020self}* & NIPS'20 & $\mathcal{O}(N^{2})$ & \gv{No} & 58.7 & 53.1 & 54.3 & \ov{48.9}\\
CCons~\cite{dai2022cluster}* & ACCV'22 & $\mathcal{O}(N^{2})$ & \gv{No} & \bv{60.3} & \bv{54.0} & \bv{57.1} & \bv{50.1} \\
%TCL~\cite{shen2023triplet} & ArXiV'23 & & & 66.3 & 60.4 & - & 63.7 & 56.2 & - & 61.1 & 52.9 & -  \\
\hline
& Speedup & \multicolumn{6}{c}{} \\ \hline
\textbf{Ours (25\%)} & $4.97\times$ & $\mathcal{O}(kN)$ & \gv{No} & 59.0 & 52.3 & 55.6 & 48.5\\
\textbf{Ours (50\%)} & $2.44\times$ &  $\mathcal{O}(kN)$ & \gv{No} & \ov{59.2} & 52.8 & 56.0 & \ov{48.9}\\
\textbf{Ours (75\%)} & $1.31\times$ &  $\mathcal{O}(kN)$ & \gv{No} & \gv{59.7} & \ov{53.3} & \ov{56.6} & \gv{49.6} \\
\textbf{Ours (100\%)} & $1.00\times$ &  $\mathcal{O}(kN)$ & \gv{No} & \gv{59.7} & \gv{53.4} & \gv{56.9} & \bv{50.1}\\
\hline
\end{tabular}
\end{table}

\begin{table}[ht]
\caption{Comparison with relevant fully-unsupervised Vehicle ReID methods in \texttt{Veri-Wild} in the medium and large setups. The best result is highlighted in \bv{blue}, the second best in \gv{green}, and the third in \ov{orange}. RRMC means Re-Ranking Memory Complexity and CPD (Cluster Parameter per Dataset) indicates if the method relies on specific clustering parameters per dataset. Speedup values are measured in comparison to the version with $100\%$ of the data. (p\%) means that p\% of all data points are sampled in LNS at each epoch.}
\label{tab:veri_wild}
\centering
\begin{tabular}{p{1.58cm}| P{1.0cm}|P{0.75cm}|P{0.4cm}|p{0.45cm}|p{0.4cm}|p{0.45cm}|p{0.4cm}}
\hline
\multicolumn{1}{c|}{} &
\multicolumn{1}{|c|}{} &
\multicolumn{1}{|c|}{} &
\multicolumn{1}{|c|}{} &
%\multicolumn{3}{|c|}{\textbf{Veri-Wild (Small)}} &
\multicolumn{2}{|c|}{\textbf{VW-Med}} &
\multicolumn{2}{|c}{\textbf{VW-Large}} \\
\hline
Method & Reference & RRMC & CPD & mAP & R1 & mAP & R1\\ \hline
BUC~\cite{lin2019bottom} & AAAI'19 &  - & \gv{No} & 14.8 & 33.8 & 9.2 & 25.2\\ 
MMCL~\cite{wang2020unsupervised} & CVPR'20 & - & \gv{No} & 19.2 & 39.1 & 14.1 & 33.1\\
SSML~\cite{yu2021unsupervised} & IROS'21 & - & \gv{No} & 20.4 & 43.9 & 15.8 & 34.7\\
%MSCL~\cite{wang2022unsupervised} & SVIP'22 & - & - & - & - & - & - & \bv{30.4} & \bv{57.2} & - \\
\hline
& Speedup & \multicolumn{6}{c}{} \\ \hline
\textbf{Ours (25\%)} & $7.60\times$ & $\mathcal{O}(kN)$ & \gv{No}  & 23.6 & 42.2 & 18.0 & 32.2\\ 
\textbf{Ours (50\%)} & $2.62\times$ & $\mathcal{O}(kN)$ & \gv{No} & \ov{25.6} & \ov{45.7} & \ov{19.8} & \ov{35.5}\\ 
\textbf{Ours (75\%)} & $1.48\times$ & $\mathcal{O}(kN)$ & \gv{No}  & \bv{26.0} & \bv{46.8} & \bv{20.3} & \bv{36.4}\\ 
\textbf{Ours (100\%)} & $1.00\times$ & $\mathcal{O}(kN)$ & \gv{No} & \gv{25.8} & \gv{46.4} & \gv{20.0} & \gv{36.2}\\ 
%\textbf{Ours (best $k$)}  & &  &  &  &  &  &  &  &  & &  & &  \\ 
\hline
\end{tabular}
\end{table}

\subsection{Results and Speedup with LNS}
\label{sec:result_speedup_local_sampling}

We evaluate our proposed LNS, for which we can take different percentages from the whole data at each iteration. %Based on the explanation of Section~\ref{sec:local_neighborhood_sampling}, we take different percentages from the whole data to feed current iterations, which means fewer data is used for feature extraction, Local Re-Ranking, Clustering, and so on. 
In Table~\ref{tab:state_of_art_reid_fully_unsupervised}, we show our results for fully-unsupervised Person ReID with different amounts of data.

With $75\%$ of the data, we obtain the second-best result for \texttt{MSMT17} against other methods that use the whole dataset. When we reduce it to $50\%$ we face a natural performance drop, but even with $25\%$ of the data, we are still on par or even better than top-tier methods. This shows our method can mine useful patterns even with a reduced amount of data.

%which increases back with $25\%$ of the data. We argue that this happens because of the random sampling process, which can sample some regions without discriminant information, which can hinder the performance. However, it is important to note that with $25\%$ of the data, we are still on par or even better than recent top-tier models that consider $100\%$ such as RTMem~\cite{yin2023real} ($60.5\%$ vs $57.1\%$ in R1), PPLR~\cite{cho2022part} ($32\%$ vs $31.4\%$ in mAP), and outperforming by a margin in all metrics CACL~\cite{li2022cluster} and HDCRL~\cite{cheng2022hybrid}. In Market, we see we have a little performance dropping for $75\%$ of the data, and since it is a relatively small dataset of $12,936$ images on training with less identities and variability (i.e., less number of cameras), the model generated degenerated clusters when we reduce it to $50\%$ or $25\%$. However, we argue it is expected since we designed our strategy to operate in large-scale training sets.

In \texttt{Veri} (Table~\ref{tab:state_of_art_vehicle}), the performance drops when considering fewer data, but the difference is small even with $50\%$ of the data. We still achieve competitive performance in comparison to previous methods using the whole dataset, and we use a more memory-efficient Re-Ranking.

In \texttt{VehicleID}, performance is stable across all percentages (Table~\ref{tab:vehicleID}). In the most challenging evaluation scenario (TS = 2400), the drop in mAP and R1 is $1.3$ and $1.6$ p.p., respectively, when going from $100\%$ to $25\%$ of the data. This is significantly lower than we observe in smaller datasets, such as \texttt{Veri} and \texttt{MSMT17}. We expect this to happen since our method was tailored to large-scale scenarios.

%our models drop performance from $100\%$ to $25\%$ of the data by $1.3$ and $1.6$ in mAP and R1 which is significantly lower than the dropping considering the smaller Veri and MSMT17. The dropping is even smaller in the other two evaluation scenarios. Moreover, our $25\%$-model is almost 5 times faster ($4.97\times$) than the $100\%$-model with affordable loss of accuracy (less than $2\%$) in the most complex evaluations scenarios. That is an important remark for large-scale fully-unlabeled datasets where a faster training and deployment is required. We also see speedups in other percentages with even less performance dropping. 

In \texttt{Veri-Wild} (Table~\ref{tab:veri_wild}), our method outperforms the prior art in most metrics with different amounts of data. With $50\%$, we perform better in all scenarios. With only $25\%$, we still outperform prior art by $2.2$ p.p. in mAP, in a harder scenario (Large), and in all metrics in the other scenarios. With this reduced amount of data, the speedup is $7.6\times$ in comparison to training with the whole dataset. With such a high speedup, %and a drop around $2.0$ p.p. in mAP, 
it can be a good trade-off for fast training and deployment. This shows our Local Neighborhood Sampling can effectively reduce the dataset size while keeping superior results in large-scale datasets. 

%The performance difference between the models trained with $100\%$ and $25\%$ are a little higher than for VehicleID, but considering a speedup of $7.60x$ and performance dropping around $2.0$ for mAP, it can be a good trade-off considering a fast training and deployment. This show our proposed Local Neighborhood Sampling can effectively reduce the training dataset size and keep high performance mainly to the large-scale datasets. 

\subsection{Visualization of Results}

\begin{figure*}[ht]
\centering
\includegraphics[width=6.8in]{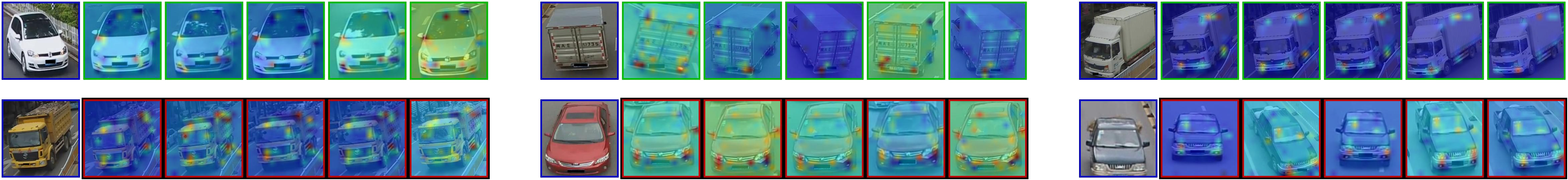}
\caption{Activation maps for the top-5 images retrieved from the gallery, given a query image (blue border) in the Veri dataset. The top row shows successful cases, and the bottom row shows failure cases. The visualizations were generated considering the ResNet50 backbone.}
\label{fig:veri_visualizations}
\end{figure*}

\begin{figure}[ht]
\centering
\includegraphics[width=2.8in]{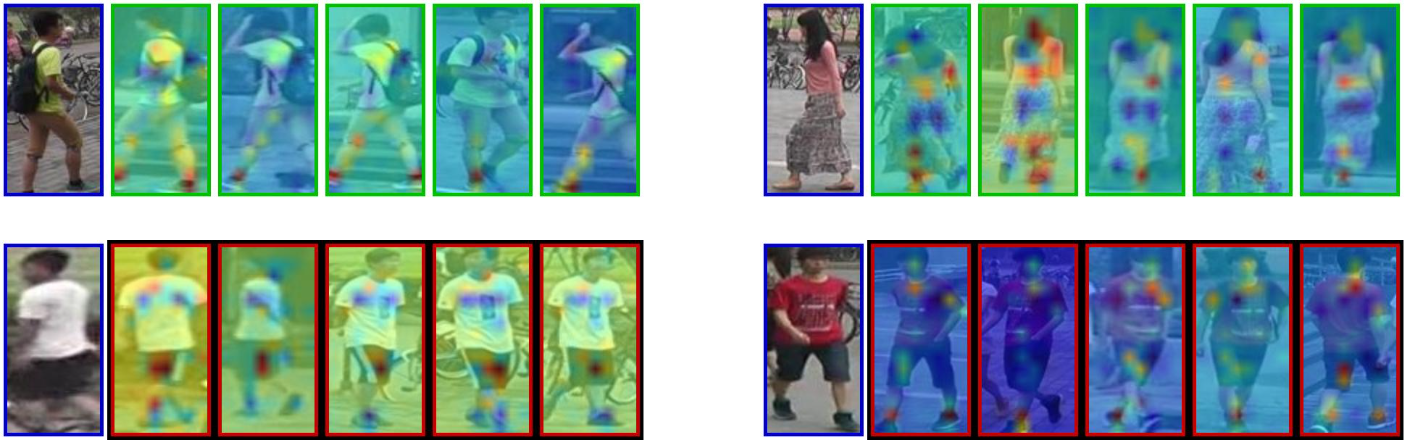}
\caption{Activation maps for the top-5 images retrieved from the gallery, given a query image (blue border) in the \texttt{Market} dataset. The top row shows successful cases, and the bottom row shows failure cases. The visualizations were generated from the ResNet50 backbone.}
\label{fig:market_visualizations}
\end{figure}

Figures~\ref{fig:veri_visualizations} and~\ref{fig:market_visualizations} depict the activated regions for the top-5 retrieved images from the gallery given a query image in \texttt{Veri} and \texttt{Market} datasets, respectively. In the top rows (correct matches), we can see that our model is able to learn fine-grained and point-of-view invariant features. In all images for both Vehicle and Person Re-Identification, just a few regions of the image are strongly activated over the identity (redder regions of the activation maps), showing that our model focuses on specific discriminant parts and is invariant to the background (no activation in any background region). In the bottom rows, we see that our model retrieves images with high visual similarity to the query. It would be hard even for a human to tell they are from different identities. % Despite these errors, our model is still able to detect fine-grained parts under different point of views and resolutions without being influenced by the background. 

\section{Ablation Study}
\label{sec:ablation_study}

We evaluate the impact of each contribution. For more controlled experimentation, when we change one factor of the method, the others remain unchanged. Experiments for the Barlow Twins initialization and the sensitivity to hyper-parameters $\tau$ and $\lambda$ are in the Supplementary Material.

\begin{table}[ht]
\caption{Ablation study. We compare the proposed Local Re-Ranking (LRR) with the Full Re-Ranking (FRR) as well as the influence of the noise-robust $\varepsilon$ scheduling (NR-$\varepsilon$) and Co-Training (CT). The \rv{\underline{ underlined}} results assume an oracle where the best $\varepsilon$ for DBSCAN is selected per dataset.}
\centering
\begin{tabular}{P{0.45cm}|P{0.1cm}|P{0.7cm}P{0.25cm}|P{0.45cm}|P{0.4cm}|P{0.45cm}|P{0.4cm}|P{0.45cm}|P{0.4cm}}
\multicolumn{4}{c}{} &
\multicolumn{2}{|c|}{\textbf{Market}} &
%\multicolumn{2}{|c}{\textbf{MSMT17}} &
\multicolumn{2}{|c}{\textbf{MSMT17}} &
\multicolumn{2}{|c}{\textbf{Veri776}} \\ \hline
& & NR-$\varepsilon$ & CT & mAP & R1 & mAP & R1 & mAP & R1 \\ \hline
\multirow{3}{*}{FRR} & \#1  &  & & \rv{\underline{86.8}} & \rv{\underline{94.4}} & \rv{\underline{42.7}} & \rv{\underline{69.7}} & \rv{\underline{39.7}} & \rv{\underline{84.6}}\\
& \#2  &\checkmark &  & 87.2 & 94.4 & 34.4 & 64.9 & 36.5 & 81.1\\ 
& \#3 &\checkmark & \checkmark & 87.8 & 94.8 & 39.1 & 68.3 & 39.5 & 84.7\\ \hline
\multirow{4}{*}{LRR} & \#4  &  & & \rv{\underline{86.4}} & \rv{\underline{94.1}} & \rv{\underline{42.5}} & \rv{\underline{69.3}} & \rv{\underline{41.6}} & \rv{\underline{85.1}} \\
& \#5 &\checkmark & & 84.8 & 93.3 & 38.8 & 67.1 & 39.8 & 82.8\\
%& \#6 & & \checkmark & \checkmark & - & - & 38.5 & 66.7 & 41.3 &  86.8\\
& \#6 & \checkmark & \checkmark & 85.8 & 94.0 & 43.2 & 70.9 & 41.3 & 86.3\\
\hline
\end{tabular}
\label{tab:ablation_study}
\end{table}

\subsection{Comparison between Local and Full Re-Ranking}

We compare our proposed LRR to FRR in terms of accuracy and time when applied in our proposed pipeline. For fairness, we run both methods on the same machine.

Table~\ref{tab:ablation_study} shows the comparison between LRR and FRR when our contributions are considered: Noise-Robust Density Scheduling (NR-$\varepsilon$), and Co-Training (CT). It is clear our LRR is even more advantageous as the complexity of the dataset increases. When NR-$\varepsilon$ and CT are not used (Line \#1 vs. line \#4 in Table~\ref{tab:ablation_study}), we assume we know the best $\varepsilon$ per dataset. In this case, for \texttt{Market} and \texttt{MSMT17}, our LRR is below FRR just by a small margin and surpasses it in the complex \texttt{Veri776} dataset. Considering all contributions (line \#3 vs. line \#6), our LRR surpasses FRR in \texttt{MSMT17} and \texttt{Veri776}.

% compare both of them in our proposed pipeline and how the performances changes when different parts are considered along with one of them. In general, we see our LRR is below FRR just by a tiny margin or surpass it in a lof of scenarios. For fairness, we compare them in our pipeline when we remove the Noise-Robust Density Scheduling (NR $\varepsilon$) and the proposed Co-Training (CT). Those results are shown in line \#1 vs line \#4. For Market and MSMT17 our LRR is below FRR just by a low margin and surpass it in Veri776, which has more images and more number of cameras than the other two datasets. Considering the whole pipeline (line \#3 vs line \#6) our LRR surpass FRR now in both MSMT17 and Veri776. Similar conclusions can be drawn from the remaining comparison lines. 

\begin{figure}[ht]
\centering
\subfloat[]{\includegraphics[width=1.5in]{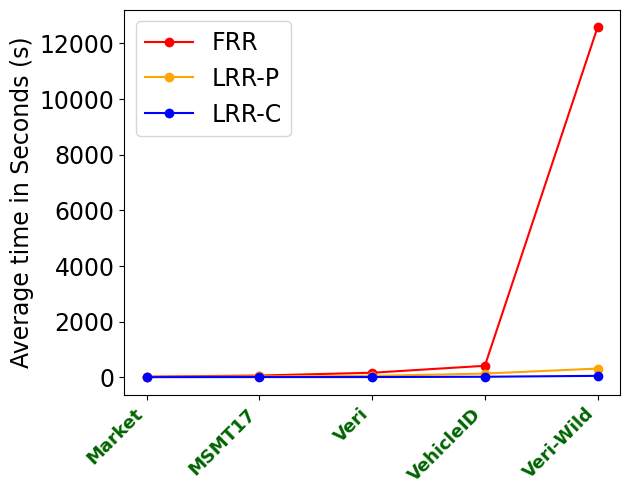}
\label{fig:FRR_vs_rest}}
\hfil
\subfloat[]{\includegraphics[width=1.5in]{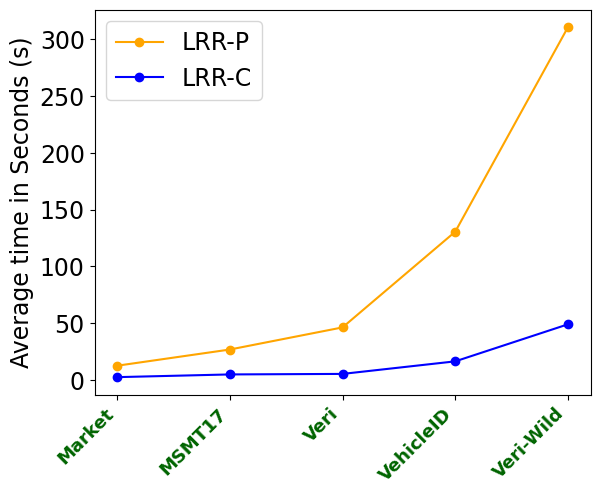}
\label{fig:LRR-P_vs_LRR-C}}
\hfil
\caption{ (a) Time comparison between Full Re-Ranking (FRR) and the proposed Local Re-Ranking (LRR) in datasets with increasing complexities. LRR is implemented in Python (LRR-P) and Cython (LRR-C). The time is an average of three runs over the three backbones. For better visualization, (b) shows only LRR-P and LRR-C.}
\label{fig:FRR_vs_LRR}
\end{figure}

We also compare re-ranking execution time when considering LRR vs. FRR (Figure~\ref{fig:FRR_vs_rest}). Corroborated by the theoretical time analysis presented in Section~\ref{subsec:local_reranking_clustering}, our LRR is faster in all datasets, which becomes more evident in the large-scale datasets \texttt{VehicleID} and \texttt{Veri-Wild}, with a Python implementation (LRR-P). LRR-P is approximately $3.14\times$ and $40\times$ faster in \texttt{VehicleID} and \texttt{Veri-Wild}, respectively. 

Targeting practical scenarios, we also developed a Cython implementation of LRR (LRR-C), which boosts, even more, the time efficiency, as shown in Figure~\ref{fig:LRR-P_vs_LRR-C}. This implementation is an improvement over our own Python implementation of LRR and is $25\times$ and $257\times$ faster than FRR in \texttt{VehicleID} and \texttt{Veri-Wild}, respectively. This analysis shows LRR is more time-efficient, theoretically and in practice, which is more suitable for large-scale scenarios.          

%Besides the accuracy, our method also show big time performances improvements as shown in Figure~\ref{fig:FRR_vs_rest}. Following the theoretical time analysis presented in Section~\ref{subsec:local_reranking_clustering}, our LRR (``LRR-P'') has lower time upper bound in all datasets and it becomes more evident in the large-scale datasets VehicleID and Veri-Wild. More specifically, our proposed LRR is approximately $3.14\times$ and $40\times$ faster in VehicleID and Veri-Wild, respectively. Considering practical scenarios, we also release our Cython implementation of LRR which boosts even more the time performance as shown in Figure~\ref{fig:LRR-P_vs_LRR-C}. With this implementation, we improve over our own Python implementation of LRR and we are $25\times$ and $257\times$ faster than the FRR in VehicleID and Veri-Wild respectively. Those results show our LRR has similar or even better performance than the well-know FRR but with a better theoretical and practical time (and memory as shown in Section~\ref{sec:comparison_state_of_art}), which is more suitable for large-scale training datasets.         

\begin{figure}[ht]
\centering
\subfloat[Market]{\includegraphics[width=1.5in]{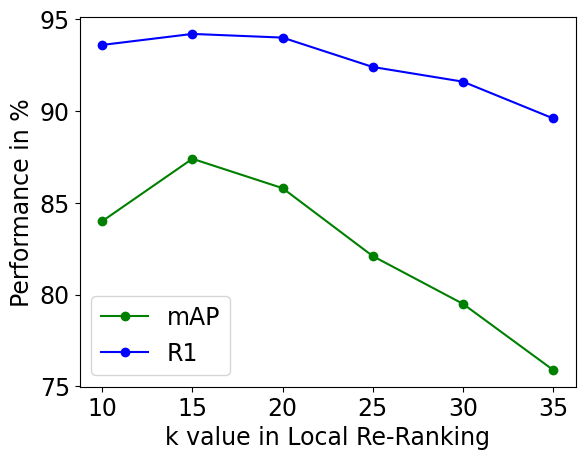}
\label{fig:Market_LRR_k}}
\hfil
\subfloat[Veri]{\includegraphics[width=1.5in]{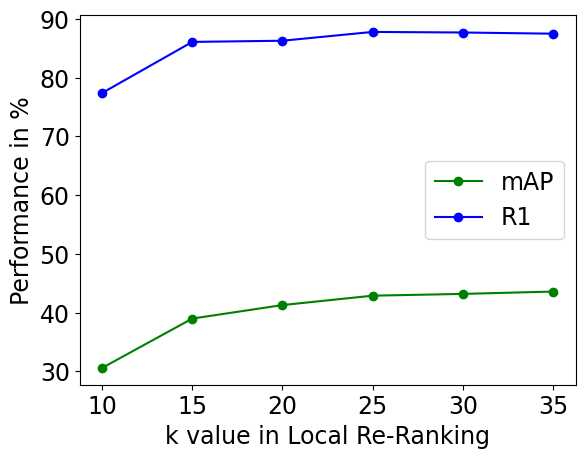}
\label{fig:Veri_LRR_k}}
\hfil
\caption{Sensibility of our model to parameter $k$ in the proposed Local Re-Ranking, for datasets (a) Market and (b) Veri.} %We always keep $k = 20$ for all experiments.}
\label{fig:LRR_k}
\end{figure}

Finally, in Figure~\ref{fig:LRR_k}, we show the impact of the nearest neighbors parameter $k$ in \texttt{Market} and \texttt{Veri}, which display contrasting behaviors when $k$ is changed. For \texttt{Market}, performance decreases for $k > 20$ but, in \texttt{Veri}, performance increases with $k$. Since our method is fully-unsupervised, we keep $k = 20$ for all other experiments and datasets.
%However, it is important to point out we have not chosen the best $k$ for each dataset, since our model operates in the fully-unsupervised scenario. So, we set $k = 20$ for all experiments.

\subsection{Impact of the Noise-Robust Density Scheduling}

\begin{figure}[ht]
\centering
\subfloat[]{\includegraphics[width=1.5in]{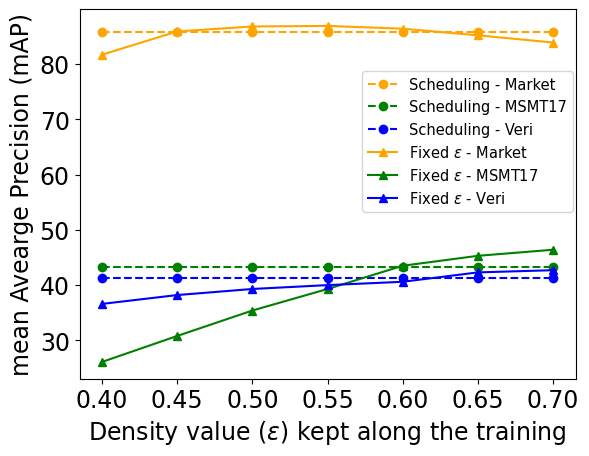}
\label{fig:mAP_eps}}
\hfil
\subfloat[]{\includegraphics[width=1.5in]{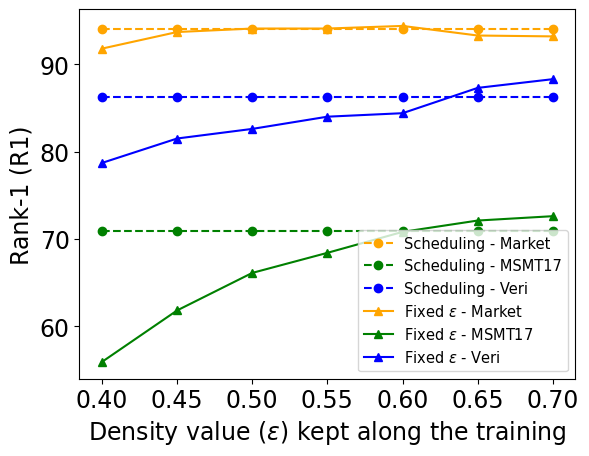}
\label{fig:R1_eps}}
\hfil
\caption{Our method's performance for three datasets, with the Noise-Robust Density Scheduling (Scheduling) and when a fixed density parameter (Fixed $\varepsilon$) is used during training, considering two metrics: (a) mAP and (b) R1.}
\label{fig:eps_sensitivity}
\end{figure}

We verify the impact of the proposed Noise-Robust Density Scheduling. Figure~\ref{fig:eps_sensitivity} shows mAP and R1 in three datasets when using our scheme and with a fixed $\varepsilon$ during training.

The performance of our proposed scheduling scheme is better than considering fixed $\varepsilon$ values for the majority of the tested values. For MSMT17 and Veri, $\varepsilon = 0.65$ or $\varepsilon = 0.7$ are optimal values; however, directly setting those values is unrealistic in fully-unsupervised scenarios. Indeed, when we compare our pipeline with and without the Noise-Robust Density Scheduling (line \#1 vs. line \#2, and line \#4 vs. line \#5 in Table~\ref{tab:ablation_study}), there is a small performance drop since our method does not have oracle knowledge about the optimal clustering parameter for each dataset. Even so, our proposed scheme does not require any grid-searching or manual selection of hyper-parameters.

%We see that our proposed scheduling has performance on pair to the ones considering a fixed $\varepsilon$ along the whole training.  For MSMT17 and Veri, $\varepsilon = 0.65$ or $\varepsilon = 0.7$ achieve the best performances, however directly set those values in our pipeline is unrealistic since our models operates in the fully-unsupervised scenario. Indeed, when we compare our pipeline with and without the Noise-Robust Density Scheduling (line \#1 vs line \#2, and line \#4 vs line \#5 in Table~\ref{tab:ablation_study}) we see a little performance dropping since our pipeline don't have any oracle knowledge about the optimal clustering parameter for each dataset. Even so, our propose scheduling is a little bit lower than those values and it does not require any grid-searching, human intervention or manual selection of the clustering threshold in the fully-unlabeled scenario. 

\begin{figure}[ht]
\centering
\subfloat[Only warmup stage]{\includegraphics[width=1.2in]{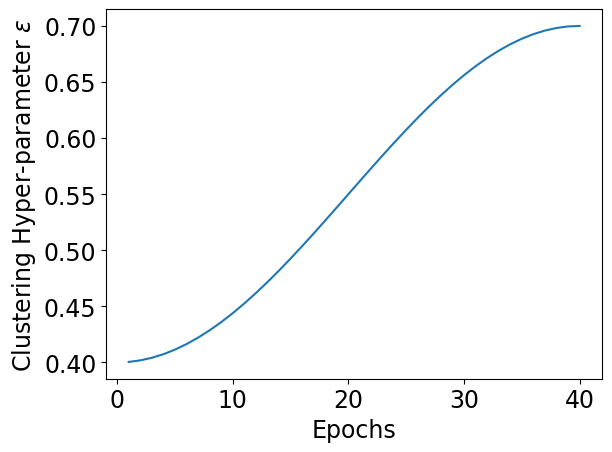}
\label{fig:just_warmup}}
\hfil
\subfloat[Only annealing stage]{\includegraphics[width=1.2in]{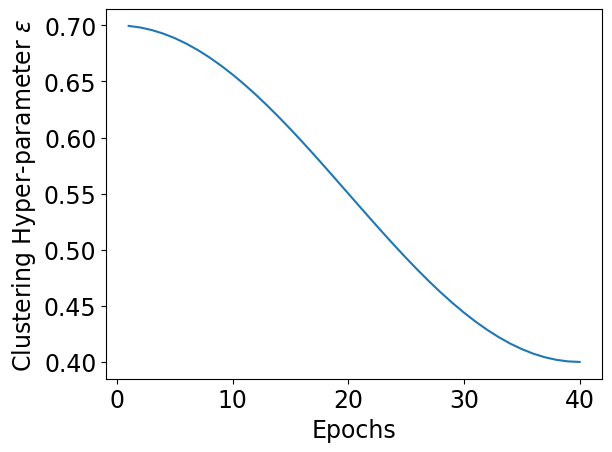}
\label{fig:just_annealing}}
\hfil
\subfloat[One cosine cycle]{\includegraphics[width=1.2in]{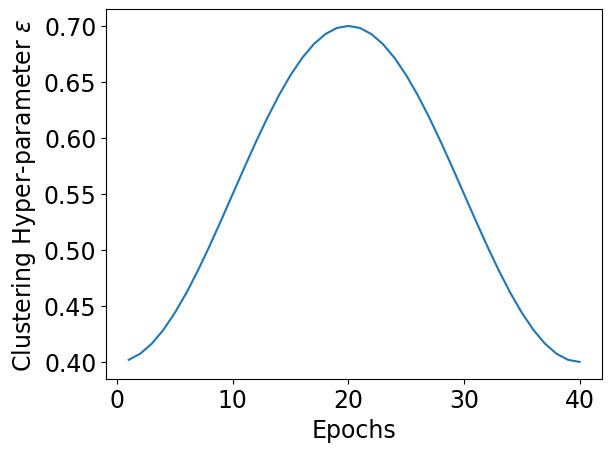}
\label{fig:full_cycle}}
\hfil
\subfloat[One + half cosine cycle]{\includegraphics[width=1.2in]{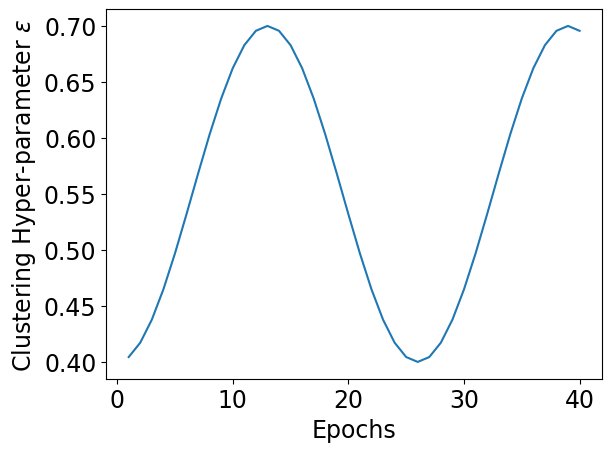}
\label{fig:full_and_half_cycle}}
\hfil
\caption{Four alternative $\varepsilon$ scheduling schemes: (a) only the warmup phase, (b) only the annealing phase, (c) one cycle of the cosine scheduling, and (d) one and half cycle of the cosine scheduling. % The performance of each one is shown in Table~\ref{tab:alternative_scheduling_results}.
}
\label{fig:alternative_schedulings}
\end{figure}

To compare our Noise-Robust Density Scheduling with other possible schemes, we test four alternatives: only the warm-up stage (Figure~\ref{fig:just_warmup}); only the annealing stage (Figure~\ref{fig:just_annealing}); and two adaptations of the cosine learning rate scheduling~\cite{loshchilov2016sgdr} but applied to the density parameter (Figures~\ref{fig:full_cycle} and~\ref{fig:full_and_half_cycle}). The results are reported in Table~\ref{tab:alternative_scheduling_results}.

%One can think if the proposed scheduling is indeed the best for the considered datasets. Since there are an unlimited number of possible schedulings, it is unfeasible to test all of them against the proposed one. For this reason, we come with four other schedulings. Two of them are based on each behavior presented in the proposed one: the first just has the ``warmup'' stage (Figure~\ref{fig:just_warmup}) and the second just the noise-robust adaptation stage (Figure~\ref{fig:just_noiserobust}). The other two are inspired in the well-known cosine learning rate scheduling proposed in~\cite{loshchilov2016sgdr}, however adapted to the density parameter scheduling in our work (Figures~\ref{fig:full_cycle} and~\ref{fig:full_and_half_cycle}). The results are depicted in Table~\ref{tab:alternative_scheduling_results}.  

\begin{table}[ht]
\caption{Our method's performance with different $\varepsilon$ scheduling schemes. The best results are in \bv{blue}.}
\centering
\begin{tabular}{P{2.8cm}|P{0.45cm}|P{0.4cm}|P{0.45cm}|P{0.4cm}|P{0.45cm}|P{0.4cm}}
\multicolumn{1}{c}{} &
\multicolumn{2}{|c|}{\textbf{Market}} &
\multicolumn{2}{|c}{\textbf{MSMT17}} &
\multicolumn{2}{|c}{\textbf{Veri776}} \\ \hline
& mAP & R1 & mAP & R1 & mAP & R1 \\ \hline
%Scheduling-1 & 85.7 & 93.8 & 46.2 & 73.1 & 41.7 & 87.1\\
%Scheduling-2 & 83.6 & 92.8 & 45.3 & 71.9 & 42.6 & 87.2\\
Only warm up & 85.3 & 93.7 & 41.8 & 70.0 & \bv{42.0} & 86.0\\
Only annealing & 86.5 & 94.2 & 35.9 & 66.1 & 39.1 & 83.7\\
One cosine cycle & 86.0 & 94.1 & 40.0 & 69.3 & 40.2 & 85.5\\
One + half cosine cycle & \bv{86.8} & \bv{94.4} & 43.1 & 70.7 & \bv{42.0} & \bv{87.0}\\ \hline
Ours & 85.8 & 94.0 & \bv{43.2} & \bv{70.9} & 41.3 & 86.3 \\\hline
\end{tabular}
\label{tab:alternative_scheduling_results}
\end{table}

Despite the fourth alternative (One + half cosine cycle) being better in \texttt{Market} and \texttt{Veri}, it is possible to verify that the cosine scheduling is sensible to the number of cycles. Our proposed scheduling outperforms the one-cycle cosine scheduling in all metrics in \texttt{MSMT17} and \texttt{Veri776}, and it is marginally inferior in \texttt{Market}. The first two strategies (Figures~\ref{fig:just_warmup} and~\ref{fig:just_annealing}) can be interpreted as part of the cosine scheduling with just a single half cycle (only \textit{warmup} or only \textit{annealing}). Our strategy performs better in most metrics in comparison to them. Therefore, to achieve the best performance with cosine scheduling, tuning the number of cycles is necessary, which is hard to do in a large-scale fully-unsupervised scenario. Our scheduling, on the other hand, does not require the number of cycles to be set. Based on the noise-robust learning theory, we designed our scheduling to be directly employed in large-scale fully-unsupervised setups.

% As expected, each part has its own merits in the training and achieves competitive performance to ours. Despite the ``Full and half cycle'' being better in Market and Veri, it has a lower performance than ours in MSMT17, and we score the second best in Veri776 dataset being below just by $0.7$ in both metrics. In counterpart the ``Full cycle`` is worse than ours in MSMT17 and Veri776. Similar conclusions we can take from the other schedulings. In general, we see that the performance difference is at most $1.0$ in all metrics in all datasets showing our competitiveness to a well-stablished scheduling.  

\subsection{Impact of Co-Training}

The impact of the proposed Co-Training strategy is shown in Table~\ref{tab:ablation_study}. When we remove the Co-Training strategy (lines \#2 and \#5) all backbones are supervised by their own generated pseudo-labels, and there is no knowledge sharing among them. We clearly see a performance drop for all metrics in all datasets. The main reason is that a backbone does not have a chance to correct itself based on the knowledge of others. In this case, any noise is propagated and amplified during training without the possibility of recovering. When co-training is in place, this problem is mitigated as discussed in Sections~\ref{sec:co_training_reid} and~\ref{sec:co_training}. Therefore, our proposed Co-Training yields performance gains for both Re-Ranking strategies without any hyper-parameter tuning or human intervention.

%Each one is optimized using their own pseudo labels, then any noise present in propagate and amplified along the epochs without any chance of correction. When they share the pseudo-labels, the noise can be mitigated as discussed in Sections~\ref{sec:co_training_reid} and~\ref{sec:co_training}. 
%Therefore, our proposed Co-Training yields performance gains for both Re-Ranking strategies, without any hyper-parameter tuning or human intervention. 

\section{Conclusions and Future Work}

We presented a novel method for fully-unsupervised Person and Vehicle Re-identification in large-scale scenarios. Most prior works rely upon costly techniques or consider unrealistic assumptions, making them infeasible for real-world deployment. For instance, they might select dataset-specific hyper-parameters, use re-ranking techniques that scale cubically or adopt ensemble methods that hinder training.

We provide contributions that enable the deployment of re-identification models in large-scale real-world applications, without the necessity for label or dataset-specific information: Local Neighborhood Sampling, Local Re-Ranking, Noise-Robust Density Scheduling, and simple Co-Training.

LNS selects a neighborhood around a random point, reducing the dataset size at each iteration. Local Re-Ranking reduces the memory and time complexities of re-ranking, by decreasing the amount of data necessary at each step, while still producing superior results. Noise-Robust Density Scheduling provides parameter-free clustering, taking into consideration the evolution of the backbones during training, which positively impacts the ability of the clustering step to deal with noisy labeling. Finally, our Co-Training technique enables inexpensive knowledge sharing among the backbones.% by simply permuting pseudo-labels sets.

Our experiments consider datasets that are often not used by other works due to their high complexity. We provide an extensive ablation study showing that our method oftentimes provides the best results with fewer assumptions. It also offers a good trade-off between execution time, memory consumption, and accuracy, especially in large-scale datasets. 

We apply our method to two different domains---person and vehicle re-identification---which are distinct in terms of the target objects and prominent features. This indicates that our contributions can be applied to other domains with similar characteristics, broadening their application.

One aspect that was only superficially explored is the number of backbones and the considered architectures. Our model relies on ResNet50 and DenseNet121 which could be replaced by lighter models, bringing even more gains in terms of memory and execution time. Another interesting exploration relates to knowledge-distilling techniques, aiming to transfer the knowledge from the ensemble to a single backbone.

As a more advanced future study, we intend to employ our solution in event understanding, leveraging its power also to produce contextual information. 
This can be done by correlating different types of targets, like people, vehicles, and places. As these targets are re-identified within an event, we can understand how they are correlated to each other and in time. This is important, for instance, in forensic investigations.   

%we will employ or solution in event understanding in the forensics scenario by aiming to find correlation between people, vehicles and places. In this way we can propose an understanding about the semantics and how the main pieces of the event are correlated to help further investigation.   

\section*{Acknowledgments}
We thank the financial support of the São Paulo Research Foundation (FAPESP), grants D\'ej\`aVu \#2017/12646-3, \#2019/15825-1, and \#2022/02299-2. We also thank the UCCS VAST Lab and Recod.ai Lab for hardware support enabling the experiments conducted in this research.


\bibliographystyle{IEEEtran}
\bibliography{IEEEabrv, refs}

%needed around the contents of the optional argument to biography to prevent
%the LaTeX parser from getting confused when it sees the complicated
% your own custom macro containing the $\backslash${\tt{includegraphics}} command to make things
%simpler here.)
 

%\bf{If you include a photo:}\vspace{-33pt}
%\begin{IEEEbiography}[{\includegraphics[width=1in,height=1.25in,clip,keepaspectratio]{fig1}}]{Michael Shell}
%Use $\backslash${\tt{begin\{IEEEbiography\}}} and then for the 1st argument use $\backslash${\tt{includegraphics}} to declare and link the author photo.
%Use the author name as the 3rd argument followed by the biography text.
%\end{IEEEbiography}



\vfill

\end{document}

\title{Supplementary Material: \\ Large-scale Fully-Unsupervised Re-Identification}

\author{Gabriel~Bertocco,
        Fernanda~Andal\'{o},~\IEEEmembership{Member,~IEEE,}
        Terrance~Boult,~\IEEEmembership{Fellow~Member,~IEEE,} and~Anderson~Rocha,~\IEEEmembership{Senior~Member,~IEEE}}
        % <-this % stops a space

\markboth{Paper submitted for a possible publication in an IEEE Transactions}%
{Bertocco \MakeLowercase{\textit{et al.}}: Large-scale Fully-Unsupervised Re-Identification}


\maketitle

Here we present further analysis and details about the method proposed in the main article. More specifically, we present a brief theoretical overview of Barlow Twins and its implementation details. We also show results considering R5 and R10 for Person Re-Identification datasets, and R5 for Vehicle Re-Identification datasets. We also provide analysis for the simpler evaluation scenarios of \texttt{VehicleID} and \texttt{Veri-Wild}. We extend our comparison to methods that use side information, such as camera labels or viewpoint information. Finally, we also show further ablation studies with Barlow Twins and the parameters $\tau$ and $\lambda$ used in the final loss function. 

\section{Theory and details on Barlow Twins}
\label{sec:barlowtwins}

In this section, we provide further explanation about Barlow Twins, the method used for the self-supervised initialization of the backbones before employing our proposed pipeline.  

Given a batch of randomly selected images $B \in X$, the algorithm generates two augmentation views for each image through some common image transformations, resulting in two new augmented batches $B^{1}$ and $B^{2}$. In our case, some image transformations might degrade Re-Identification performance, so we consider just random cropping, horizontal flipping, and shifts in brightness, contrast, and saturation. After that, we feed both batches to the backbone $f$ and get the feature representations $Z^{1}, Z^{2} \in R^{N \times d}$, respectively, for each batch, where $d$ is the feature dimension. The features are normalized by mean subtraction and diving by the standard deviation per dimension. After that, we train the model to achieve maximum decorrelation between the dimensions to encourage the model to learn complementary features and, at the same time, be robust among different augmentations. To do so, the algorithm performs the multiplication of batch features, $C = (Z^{1})^{T}Z^{2} \in R^{d \times d}$, to calculate the cross-correlation among the different dimensions of the feature vectors. The next step is to maximize the agreement between features in the same dimension (represented by values in the main diagonal of $C$) and minimize it between features from different dimensions through the following loss function:
\begin{equation}
\label{eq:barlow_twins_equation}
L_{BT} = \sum_{i}(1-C_{ii})^2 + \lambda_{BT}\sum_{i}\sum_{j\neq i}C_{ij}^2,
\end{equation}
\noindent where the first term aims to increase the invariance representation between different augmentation views of the same image, and the second decouples feature representation considering the dimensions; and $\lambda_{BT}$ weights the contribution of the second term in the loss. For further details, we refer to the original article~\cite{zbontar2021barlow}.

%The labels $Y(f,D,\varepsilon) = \{y_{i}\}_{i=1}^{N}$ also depend on the backbone $f$, so we consider three different backbones ($f_{1}, f_{2}, f_{3}$) and perform self-supervised pre-training in each one is initialized independently by doing three epochs of pre-training, then we fine-tuned them with the proposed pipeline. We are the first to proposed a Barlow Twins-based initialization for Person Re-Identification considering three different backbones before employing the proposed unsupervised learning pipeline. 

\section{Implementation Details}

We perform the self-supervised initialization with Barlow Twins for three epochs. The setup for each experiment may vary due to the availability of GPUs and memory constraints, as the datasets have different sizes. For \texttt{MSMT17}, \texttt{Veri776}, and \texttt{Veri-Wild}, we used five NVIDIA RTX A6000 with 49GB of RAM, and a batch size of 1024. For \texttt{Market} and \texttt{VehicleID}, we keep the same setup, except that for \texttt{Market}, we train OSNet in three Quadro RTX 8000 with a batch size of 768, and for \texttt{VehicleID}, we set the batch size to 768 for OSNet and DenseNet121. The Adam~\cite{kingma2014adam} optimizer was employed with a learning rate set to $3.5e-5$ for the Person ReID datasets and $3.5e-6$ for Vehicle ReID datasets. The weight decay was set to $1.5e-6$, and the $\lambda_{BT}$ in Equation~\ref{eq:barlow_twins_equation} was set to $5e-3$ for all datasets following the original implementation~\cite{zbontar2021barlow}. 

\section{Impact of Pre-training with Barlow Twins}

The impact of employing the self-supervised pre-training with Barlow Twins (BT) is shown in Table~\ref{tab:ablation_study}. We compare this initialization against simple pre-training with ImageNet. Pre-training with BT has proved crucial. For \texttt{Market}, our method produces state-of-the-art results when the self-supervised initialization is used, but yields degenerated clusters in the first epochs when BT is not used. For \texttt{Veri776}, results are similar in both scenarios; however, for \texttt{MSMT17}, mAP and R1 increase by $4.7$ and $4.2$, respectively, when BT is considered. 

\begin{table}[ht]
\caption{Comparison between our model with and without pre-training with Barlow Twins (BT)}
\centering
\begin{tabular}{P{0.1cm}|P{0.25cm}|P{0.45cm}|P{0.4cm}|P{0.45cm}|P{0.4cm}|P{0.45cm}|P{0.4cm}}
\multicolumn{2}{c}{} &
\multicolumn{2}{|c|}{\textbf{Market}} &
\multicolumn{2}{|c}{\textbf{MSMT17}} &
\multicolumn{2}{|c}{\textbf{Veri776}} \\ \hline
& BT & mAP & R1 & mAP & R1 & mAP & R1 \\ \hline
%\multirow{4}{*}{LRR} & \#4 & \checkmark  &  & & \rv{\underline{86.4}} & \rv{\underline{94.1}} & \rv{\underline{42.5}} & \rv{\underline{69.3}} & \rv{\underline{41.6}} & \rv{\underline{85.1}} \\
%& \#5 &\checkmark &\checkmark & & 84.8 & 93.3 & 38.8 & 67.1 & 39.8 & 82.8\\
\#1 & & - & - & 38.5 & 66.7 & 41.3 &  86.8\\
\#2 & \checkmark & 85.8 & 94.0 & 43.2 & 70.9 & 41.3 & 86.3\\
\hline
\end{tabular}
\label{tab:ablation_study}
\end{table}

\section{Impact of loss hyper-parameters}

In this section, we verify the impact of the hyper-parameters $\tau$ and $\lambda$ in the loss function and its terms (Equations 9, 10, and 11 in the main article).
%\begin{equation}
%\label{eq:final_loss}
%   L_{final} = L_{proxy} + \lambda L_{hard}.
%\end{equation}

\begin{figure}[ht]
\centering
\subfloat[]{\includegraphics[width=1.6in]{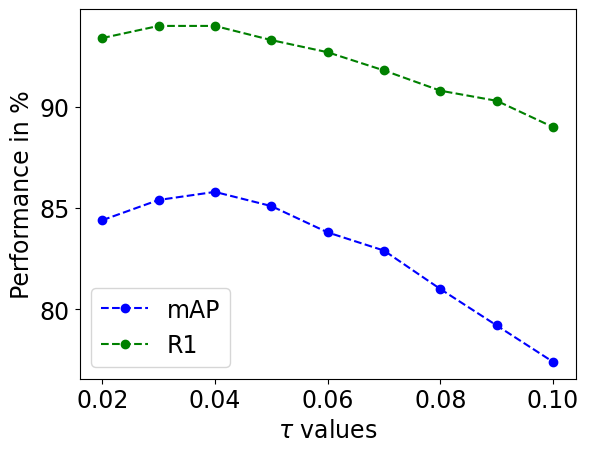}
\label{fig:tau_Market}}
\hfil
\subfloat[]{\includegraphics[width=1.6in]{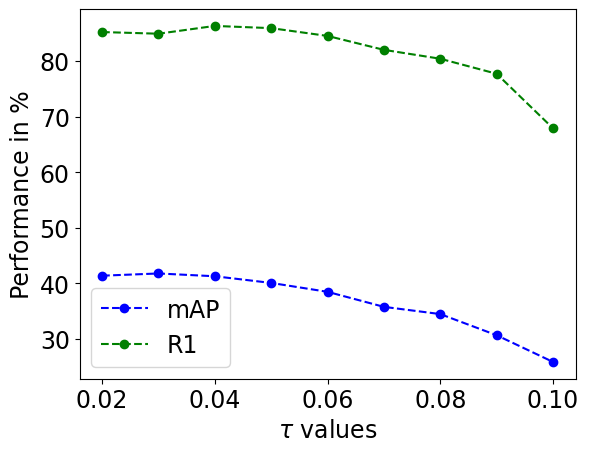}
\label{fig:tau_Veri}}
\hfil
\caption{mAP and R1 variation for different values of the $\tau$ parameter in the loss function, considering the datasets (a) \texttt{Market} and (b) \texttt{Veri776}.}
\label{fig:tau_sensitivity}
\end{figure}

The $\tau$ parameter has the goal of changing the distribution of the scores, which allows smoother gradients to aid the optimization. Its impact is verified in Figure~\ref{fig:tau_sensitivity} for the \texttt{Market} and \texttt{Veri776} datasets. We see that mAP and R1 reach their maximum when $\tau = 0.04$. After that, performance starts to decrease rapidly, for both datasets. This happens because the gradients increase together with  $\tau$, causing instability during training. Therefore, we set $\tau = 0.04$ for all datasets.

\begin{figure}[ht]
\centering
\subfloat[]{\includegraphics[width=1.6in]{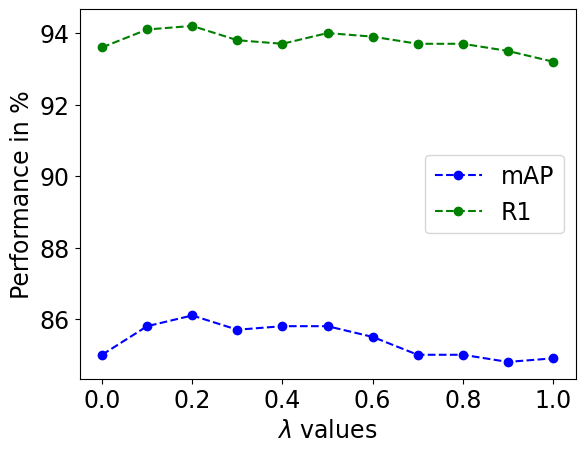}
\label{fig:lambda_Market}}
\hfil
\subfloat[]{\includegraphics[width=1.6in]{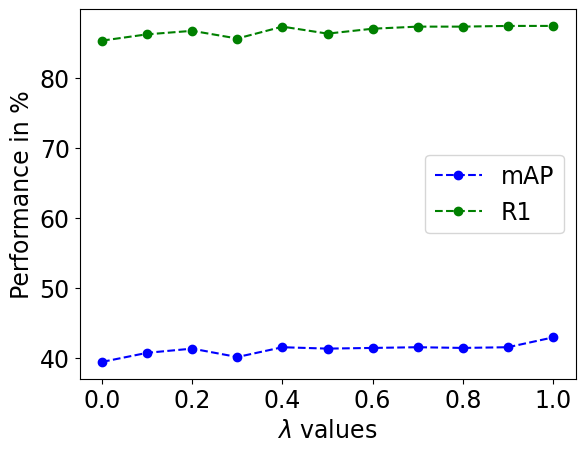}
\label{fig:lambda_Veri}}
\hfil
\caption{mAP and R1 variation for different values of the $\lambda$ parameter in the loss function, considering the datasets (a) \texttt{Market} and (b) \texttt{Veri776}.}
\label{fig:lambda_sensitivity}
\end{figure}

The $\lambda$ value weights the contribution of the batch-hard triplet loss ($L_{hard}$) in the final loss function. While $L_{proxy}$ is a more global loss term since it considers all class proxies for optimization, $L_{hard}$ enforces a more local view since the hard triplets are mined at the batch level. That is, $\lambda = 0.0$ means that there is no local contribution, while a too-large value might make $L_{hard}$ dominant over $L_{proxy}$ and hinder model optimization. The impact of the $\lambda$ value is shown in Figure~\ref{fig:lambda_sensitivity}.

When $\lambda = 0.0$ (i.e., no $L_{hard}$), the performance is among the worst for both datasets, which shows the importance of having a local view during optimization. Higher values negatively impact the performance in the \texttt{Market} dataset, but it does not affect results in \texttt{Veri776}. To achieve a trade-off we keep $\lambda = 0.5$ for all datasets. 

\section{Extended Results}

% \subsection{Extended Re-Identification Results}

We extend the comparison from the main article by including R5 and R10 for Person Re-Identification and R5 for Vehicle Re-Identification (Tables~\ref{tab:state_of_art_reid_fully_unsupervised}, \ref{tab:vehicleID_fully_unsupervised}, and~\ref{tab:veri_wild_fully_unsupervised}). For \texttt{VehicleID} and \texttt{Veri-Wild}, we also show the performance in the simpler evaluation scenarios (the medium and largest scenarios are presented in the main paper). 

For Person ReID, the results in Table~\ref{tab:state_of_art_reid_fully_unsupervised} follow the same conclusions presented in the main article. We have the best performance for all metrics in \texttt{MSMT17}, and our method, with only $75\%$ of the data, still ranks second place in this dataset. For \texttt{Market}, we show competitive results, ranking third place in R5 and being below AdaMG in R10 just by $0.4$ p.p. However, AdaMG adopts an unrealistic scenario where the clustering hyper-parameter is tuned per dataset. Moreover, AdaMG adopts a more memory- and time-complex Re-Ranking strategy. Therefore, with a less complex method and without requiring any per-dataset hyper-parameter tuning, we achieve competitive performance in \texttt{Market} and outperform prior art in all metrics in \texttt{MSMT17} with $100\%$ and $75\%$ of the data.      

\begin{table*}[ht]
%\vspace{-55pt}
\caption{Comparison with relevant fully-unsupervised Person ReID methods. The best result is highlighted in \bv{blue}, the second best in \gv{green}, and the third in \ov{orange}. RRMC means Re-Ranking Memory Complexity and CPD (Cluster Parameter per Dataset) indicates if the method relies on specific clustering parameters per dataset. (p\%) means that p\% of all data points are sampled in the Local Neighborhood Sampling and used in the current epoch.}
\label{tab:state_of_art_reid_fully_unsupervised}
\centering
\begin{tabular}{|p{2.4cm}| p{1.2cm}|P{1.3cm}|P{0.8cm}|p{0.7cm}|p{0.7cm}|p{0.7cm}|p{0.7cm}|p{0.7cm}|p{0.7cm}|p{0.7cm}|p{0.7cm}|}
\hline
\multicolumn{1}{|c|}{} &
\multicolumn{1}{|c|}{} &
\multicolumn{1}{|c|}{} &
\multicolumn{1}{|c|}{} &
\multicolumn{4}{|c|}{\textbf{Market}} &
\multicolumn{4}{|c|}{\textbf{MSMT17}} \\ \hline
%\multicolumn{14}{|c|}{\textit{Fully Unsupervised}}  \\ \hline
Method & Reference & RRMC & CPD & mAP & R1 & R5 & R10
& mAP & R1 & R5 & R10 \\ \hline
ABMT~\cite{chen2020enhancing} & WACV'20 & $\mathcal{O}(N^{2})$ & \gv{No} & 65.1 & 82.6 & - & - & - & - & - & - \\
SpCL~\cite{ge2020self} & NeurIPS'20 & $\mathcal{O}(N^{2})$ & \gv{No} & 73.1 & 88.1 & 95.1 & 97.0 & 19.1 & 42.3 & 55.6 & 61.2 \\
GCL+~\cite{chen2022learning} & TPAMI'22 & $\mathcal{O}(N^{2})$ & \gv{No} & 69.3 & 89.0 & 94.6 & 96.0 & 22.0 & 47.9 & 61.3 & 67.1 \\ 
GSam~\cite{han2022rethinking} & TIP'22 & $\mathcal{O}(N^{2})$ & \gv{No} & 79.2 & 92.3 & 96.6 & 97.8 & 24.6 & 56.2 & 67.3 & 71.5 \\
%SpCL-IBN~\cite{ge2020self} & NeurIPS'20 & 73.8 & 88.4 & 95.3 & 97.3 & - & - & - & - & 24.0 & 48.9 & 61.8 & 67.1 \\
RLCC~\cite{zhang2021refining} & CVPR'21 & $\mathcal{O}(N^{2})$ & \gv{No}&  77.7 & 90.8 & 96.3 & 97.5 & 27.9 & 56.5 & 68.4 & 73.1 \\
%MaskPre~\cite{yin2022unsupervised} & PR'22 & - & \gv{No} & 77.7 & 90.4 & 96.4 & 97.9 & 26.7 & 58.5 & 69.3 & 73.8 \\
HCLP~\cite{zheng2021online} & ICCV'21 & - & \gv{No} & 78.1 & 91.1 & 96.4 & 97.7 & 26.9 & 53.7 & 65.3 & 70.2 \\
%DFC~\cite{si2023diversity} & IPM'23 & - & \gv{No} &  78.6 & 90.8 & 97.5 & - & - & - & - & - \\
ICE~\cite{Chen_2021_ICCV} & ICCV'21 & $\mathcal{O}(N^{2})$ & \gv{No} &79.5 & 92.0 & 97.0 & 98.1 & 29.8 & 59.0 & 71.7 & 77.0 \\
%MGCE-HCL~\cite{sun2022hybrid} & ACPR'22 & - & \gv{No} & 79.6 & 92.1 & - & - & - & - & - & - \\
%MPC~\cite{li2023multi} & CVIU'23 &$\mathcal{O}(|C_{i}|^{2})$ & \gv{No} & 77.4 & 90.9 & 96.4 & 97.6 & - & - & - & - \\
CACL~\cite{li2022cluster} & TIP'22 & - & \gv{No} & 80.9 & 92.7 & 97.4 & 98.5 & 23.0 & 48.9 & 61.2 & 66.4 \\
HDCRL~\cite{cheng2022hybrid} & TIP'22 & - & \gv{No} & 81.7 & 92.4 & 97.4 & 98.1 & 24.6 & 50.2 & 61.4 & 65.7 \\
PPLR~\cite{cho2022part} & CVPR'22 & $\mathcal{O}(N^{2})$ & \gv{No} & 81.5 & 92.8 & 97.1 & 98.1 & 31.4 & 61.1 & 73.4 & 77.8 \\
%RMCL~\cite{pang2023reliability} & KBS'23 & $\mathcal{O}(N^{2})$ & \gv{No} & 81.7 & 93.0 & 97.6 & 98.4 & 32.5 & 62.3 & 73.6 & 78.0 \\
%Zhang et. al.~\cite{zhang2022unsupervised} & AI'22 & $\mathcal{O}(N^{2})$ & \gv{No} & 83.0 & 93.2 & 97.3 & - & - & - & - & - \\
RTMem~\cite{yin2023real} & TIP'23 & - & \rv{Yes} & 83.0 & 92.8 & 97.4 & 98.3 & 32.8 & 57.1 & 70.0 & 74.9\\
CCons~\cite{dai2022cluster} & ACCV'22 & $\mathcal{O}(N^{2})$ & \gv{No} & 83.0 & 92.9 & 97.2 & 98.0 & 33.0 & 62.0 & 71.8 & 76.7 \\
%MCL~\cite{jin2022meta} & ICM'22 & $\mathcal{O}(N^{2})$ & \rv{Yes} & 83.3 & 93.0 & - & - & 33.4 & 62.9 & - & - \\
%GCM~\cite{zhang2022graph} & SIVP'22 & $\mathcal{O}(N^{2})$ & \gv{No} & 83.4 & 93.3 & - & - & - & - & - & - \\
ISE~\cite{zhang2022implicit} & CVPR'22 & - & \rv{Yes} & \gv{84.7} & \bv{94.0} & \gv{97.8} & \gv{98.8} & 35.0 & 64.7 & 75.5 & 79.4 \\
%Zheng~\cite{zheng2022multi} & ArXiv'22 & $\mathcal{O}(N^{2})$ & \gv{No} & 83.1 & 92.8 & 97.1 & 98.0 & 35.6 & 63.8 & 75.3 & 79.5 \\
HHCL~\cite{hu2021hard} & NIDC'21 & $\mathcal{O}(N^{2})$ & \gv{No} &  84.2 & \ov{93.4} & \ov{97.7} & 98.5 & - & - & - & - \\
GRACL~\cite{zhang2022global} & TCSVT'22 & $\mathcal{O}(N^{2})$ & \gv{No} &  83.7 & 93.2 & 97.6 & \ov{98.6} & 34.6 & 64.0 & 75.0 & 79.3 \\
%CGCL~\cite{miao2022confidence} & ArXiv'22 & - & \gv{No} & 85.3 & 94.2 & 97.6 & 98.5 & 34.6 & 63.4 & 74.6 & 79.3 \\
%NCPLR~\cite{cheng2022neighbour} & ArXiv'22 & - & \rv{Yes} & 86.3 & 94.3 & 98.0 & 98.7 & 35.7 & 66.3 & 76.9 & 80.6 \\
AdaMG~\cite{peng2023adaptive} & TCSVT'23 &  $\mathcal{O}(N^{2})$ & \rv{Yes} & \ov{84.6} & \gv{93.9} & \bv{97.9} & \bv{98.9} & \ov{38.0} & \ov{66.3} & \ov{76.9} & \ov{80.6} \\
%DCCT~\cite{chen2022dual} & ArXiv'22 &  $\mathcal{O}(N^{2})$ & \rv{Yes} & 86.3 & 94.4 & 97.7 & 98.5 & 41.8 & 68.7 & 79.0 & 82.6 \\ 
\hline
%\textbf{Ours} & & 84.4 & 93.2 & 96.7 & 97.6 & 45.3 & 71.9 & 81.0 & 83.9\\
%\textbf{Ours (best $k$)} & & 85.7 & 93.9 & 97.2 & 98.0 &  &  &  & \\
\textbf{Ours (25\%)} &  & $\mathcal{O}(kN)$& \gv{No} & - & - & - & - & 32.0 & 60.5 & 71.0 & 75.2\\
\textbf{Ours (50\%)} & & $\mathcal{O}(kN)$ & \gv{No} & - & - & - & - & 24.3 & 50.4 & 60.6 & 65.4\\
\textbf{Ours (75\%)} & & $\mathcal{O}(kN)$ & \gv{No} & 82.9 & 92.6 & 97.0 & 97.8 & \gv{39.3} & \gv{67.3} & \gv{77.3} & \gv{80.8}\\
\textbf{Ours (100\%)} &  & $\mathcal{O}(kN)$ & \gv{No} & \bv{85.8} & \bv{94.0} & \ov{97.7} & 98.5 & \bv{43.2} & \bv{70.9} & \bv{80.8} & \bv{84.2}\\

\hline
\end{tabular}
\end{table*}

In \texttt{VehicleID} (Table~\ref{tab:vehicleID_fully_unsupervised}), our method presents the same behavior shown in the main article for the simplest evaluation scenario (``Test size = 800''). That is, we obtain the best or second-best performance in all metrics. The same happens for R5 in ``Test size = 1600'' and ``Test size = 2400''. 

\begin{table*}[ht]
\caption{Comparison with relevant fully-unsupervised Vehicle ReID methods in \texttt{VehicleID}. The best result is highlighted in \bv{blue}, the second best in \gv{green}, and the third in \ov{orange}. RRMC means Re-Ranking Memory Complexity and CPD (Cluster Parameter per Dataset) indicates if the method relies on specific clustering parameters per dataset. Methods with * were reproduced from~\cite{he2022multi}, which seems to follow the same evaluation protocol as ours. (p\%) means that p\% of all data points are sampled in the Local Neighborhood Sampling and used in the current epoch.}
\label{tab:vehicleID_fully_unsupervised}
\centering
\begin{tabular}{p{2.7cm}| P{1.5cm}|P{1.0cm}|P{0.7cm}|p{0.7cm}|p{0.5cm}|p{0.5cm}|p{0.5cm}|p{0.5cm}|p{0.5cm}|p{0.5cm}|p{0.5cm}|p{0.5cm}}
\hline
\multicolumn{1}{c|}{} &
\multicolumn{1}{|c|}{} &
\multicolumn{1}{|c|}{} &
\multicolumn{1}{|c|}{} &
\multicolumn{3}{|c|}{\textbf{Test size  = 800}} &
\multicolumn{3}{|c|}{\textbf{Test size  = 1600}} &
\multicolumn{3}{|c}{\textbf{Test size  = 2400}} \\
\hline
%\multicolumn{14}{|c|}{\textit{Camera-based}} \\ \hline
%\multicolumn{14}{|c|}{\textit{Segmentation-based}} \\ \hline
%\multicolumn{13}{|c|}{\textit{Part-based Models}} \\ \hline
%Method & Reference & RR-MC & CPD & mAP & R1 & R5 & mAP & R1 & %R5 & mAP & R1 & R5 \\ \hline
%MLPL~\cite{he2022multi}* & TVT'22 & - & \gv{No} & 65.3 & 61.1 & %69.8 & 62.7 & 57.3 & 68.4 & 59.6 & 52.4 & 66.2 \\ \hline
%\multicolumn{13}{|c|}{\textit{Fully Unsupervised}} \\ \hline
Method & Reference & RRMC & CPD &  mAP & R1 & R5 & mAP & R1 & R5 & mAP & R1 & R5 \\ \hline
BUC~\cite{lin2019bottom}* & AAAI'19 & - & \gv{No} & 51.8 & 49.5 & 62.6 & 46.2 & 45.9 & 59.8 & 42.4 & 39.7 & 57.3 \\
%PAL~\cite{peng2020unsupervised}* & ArXiv'20 & - & \gv{No} & 53.5 & 50.3 & 64.9 & 48.1 & 44.3 & 60.9 & 45.1 & 41.1 & 59.1 \\ 
MAC~\cite{zhu2022manifold} & KBS'22 & - & \gv{No} & 56.2 & 54.3 & 71.1 & 51.9 & 47.5 & 66.8 & 47.4 & 44.4 & \bv{65.9} \\
%MSCL~\cite{wang2022unsupervised} & SVIP'22 & $\mathcal{O}(N^{2})$ & \gv{No} &  - & - & - & - & - & - & 52.9 & \bv{51.8} & - \\
SpCL~\cite{ge2020self}* & NeurIPS'20 & $\mathcal{O}(N^{2})$ & \gv{No} &  60.2 & 55.4 & 67.5 & 58.7 & 53.1 & 67.1 & 54.3 & \ov{48.9} & 64.8 \\
CCons~\cite{dai2022cluster}* & ACCV'22 & $\mathcal{O}(N^{2})$ & \gv{No} & \bv{62.6} & \bv{57.7} & \ov{68.0} & \bv{60.3} & \bv{54.0} & \bv{67.9} & \bv{57.1} & \bv{50.1} & \bv{65.9} \\
%TCL~\cite{shen2023triplet} & ArXiV'23 & & & 66.3 & 60.4 & - & 63.7 & 56.2 & - & 61.1 & 52.9 & -  \\
\hline
& Speedup & \multicolumn{11}{c}{} \\ \hline
\textbf{Ours (25\%)} & $4.97\times$ & $\mathcal{O}(kN)$ & \gv{No} & 61.0 & 55.1 & 68.0 & 59.0 & 52.3 & 67.2 & 55.6 & 48.5 & 64.3 \\
\textbf{Ours (50\%)} & $2.44\times$ &  $\mathcal{O}(kN)$ & \gv{No} & 61.0 & 55.0 & \gv{68.1} & \ov{59.2} & 52.8 & 67.1 & 56.0 & \ov{48.9} & 64.5 \\
\textbf{Ours (75\%)} & $1.31\times$ &  $\mathcal{O}(kN)$ & \gv{No} & \ov{61.6} & \ov{55.7} & \bv{68.6} & \gv{59.7} & \ov{53.3} & \ov{67.6} & \ov{56.6} & \gv{49.6} & \ov{65.0} \\
\textbf{Ours (100\%)} & $1.00\times$ &  $\mathcal{O}(kN)$ & \gv{No} & \gv{61.7} & \gv{56.0} & \bv{68.6} & \gv{59.7} & \gv{53.4} & \gv{67.7} & \gv{56.9} & \bv{50.1} & \gv{65.2}\\
\hline
\end{tabular}
\end{table*}

Similarly to the previous datasets, the conclusion for \texttt{Veri-Wild}, shown in Table~\ref{tab:veri_wild_fully_unsupervised}, follows the same ones in the main article. For \texttt{Veri-Wild (Small)}, we have the best performance with $75\%$ of the data, the second-best with $100\%$ of the data, and the third-best with $50\%$ of the data. This shows that even on simpler evaluation scenarios and higher ranking metrics (e.g., R5 and R10), our method still provides competitive or, as in most cases, state-of-the-art performance compared to the prior art.  

\begin{table*}[ht]
\caption{Comparison with relevant fully-unsupervised Vehicle ReID methods in \texttt{Veri-Wild}. The best result is highlighted in \bv{blue}, the second best in \gv{green}, and the third in \ov{orange}. RRMC means Re-Ranking Memory Complexity and CPD (Cluster Parameter per Dataset) indicates if the method relies on specific clustering parameters per dataset. Speedup values are measured in comparison to the version with $100\%$ of the data. (p\%) means that p\% of all data points are sampled in the Local Neighborhood Sampling and used in the current epoch.}
\label{tab:veri_wild_fully_unsupervised}
\centering
\begin{tabular}{p{2.7cm}| P{1.2cm}|P{1.0cm}|P{0.7cm}|p{0.5cm}|p{0.5cm}|p{0.5cm}|p{0.5cm}|p{0.5cm}|p{0.5cm}|p{0.5cm}|p{0.5cm}|p{0.5cm}}
\hline
\multicolumn{1}{c|}{} &
\multicolumn{1}{|c|}{} &
\multicolumn{1}{|c|}{} &
\multicolumn{1}{|c|}{} &
\multicolumn{3}{|c|}{\textbf{Veri-Wild (Small)}} &
\multicolumn{3}{|c|}{\textbf{Veri-Wild (Medium)}} &
\multicolumn{3}{|c}{\textbf{Veri-Wild (Large)}} \\
\hline
Method & Reference & RR-MC & CPD & mAP & R1 & R5 & mAP & R1 & R5 & mAP & R1 & R5 \\ \hline
BUC~\cite{lin2019bottom} & AAAI'19 &  - & \gv{No} & 15.2 & 37.5 & 53.0 & 14.8 & 33.8 & 51.1 & 9.2 & 25.2 & 41.6 \\ 
MMCL~\cite{wang2020unsupervised} & CVPR'20 & - & \gv{No} & 15.9 & 40.1 & 63.5 & 19.2 & 39.1 & 60.4 & 14.1 & 33.1 & 50.4 \\
SSML~\cite{yu2021unsupervised} & IROS'21 & - & \gv{No} & 23.7 & 49.6 & 71.0 & 20.4 & 43.9 & 64.9 & 15.8 & 34.7 & 55.4 \\
%MSCL~\cite{wang2022unsupervised} & SVIP'22 & - & - & - & - & - & - & \bv{30.4} & \bv{57.2} & - \\
\hline
& Speedup & \multicolumn{11}{c}{} \\ \hline
\textbf{Ours (25\%)} & $7.60\times$ & $\mathcal{O}(kN)$ & \gv{No}  & 28.0 & 50.4 & 74.4 & 23.6 & 42.2 & 66.5 & 18.0 & 32.2 & 55.1\\ 
\textbf{Ours (50\%)} & $2.62\times$ & $\mathcal{O}(kN)$ & \gv{No} & \ov{29.8} & \ov{53.5} & \ov{76.4} & \ov{25.6} & \ov{45.7} & \ov{69.2} & \ov{19.8} & \ov{35.5} & \ov{58.6}\\ 
\textbf{Ours (75\%)} & $1.48\times$ & $\mathcal{O}(kN)$ & \gv{No}  & \bv{30.2} & \bv{54.6} & \bv{77.1} & \bv{26.0} & \bv{46.8} & \bv{70.0} & \bv{20.3} & \bv{36.4} & \bv{59.2}\\ 
\textbf{Ours (100\%)} & $1.00\times$ & $\mathcal{O}(kN)$ & \gv{No}  & \gv{29.9} & \gv{54.1} & \gv{76.6} & \gv{25.8} & \gv{46.4} & \gv{69.2} & \gv{20.0} & \gv{36.2} & \gv{59.0}\\ 
%\textbf{Ours (best $k$)}  & &  &  &  &  &  &  &  &  & &  & &  \\ 
\hline
\end{tabular}
\end{table*}

% \subsection{Comparison to Not-Fully-Unsupervised Re-Identification methods}

We also compare, in Table~\ref{tab:state_of_art_reid}, our method to not-fully unsupervised re-identification methods, i.e., the ones that leverage camera labels or tracklets to help the optimization. Particularly, the camera labels provide strong regularization since they enable the model to train in the same scenario of the cross-camera evaluation. For this reason, methods that consider camera labels have usually higher performance. However, they are \textbf{not} fully unsupervised.

% In Table~\ref{tab:state_of_art_reid}, we provide further results comparing our method to our prior works that leverage our camera labeling or tracklets to help model learning. Particularly, the camera labeling brings a strong regularization performance since it enables the model to train in the same scenario of the cross-camera evaluation. However, the camera labeling might not be available, and the models are not fully unsupervised. For this reason, usually, models that consider camera labels have higher performance. 

\begin{table*}[ht]
%\vspace{-55pt}
\caption{Comparison with relevant Person ReID methods considering some meta-information or camera/viewpoint labeling. The best result is highlighted in \bv{blue}.}
\label{tab:state_of_art_reid}
\centering
\begin{tabular}{|p{2.5cm}| p{1.8cm}|p{0.7cm}|p{0.7cm}|p{0.7cm}|p{0.7cm}|p{0.7cm}|p{0.7cm}|p{0.7cm}|p{0.7cm}|}
\hline
\multicolumn{1}{|c|}{} &
\multicolumn{1}{|c|}{} &
\multicolumn{4}{|c|}{\textbf{Market}} &
\multicolumn{4}{|c|}{\textbf{MSMT17}} \\
\hline
Method & Reference & mAP & R1 & R5 & R10
& mAP & R1 & R5 & R10 \\ \hline
\multicolumn{10}{|c|}{\textit{Camera-based}} \\ \hline
SSL~\cite{lin2020unsupervised} & CVPR'20 & 37.8 & 71.7 & 83.8 & 87.4 & - & - & - & - \\
CCSE~\cite{lin2020unsupervisedccse} & TIP'20 & 38.0 & 73.7 & 84.0 & 87.9 & 9.9 & 31.4 & 41.4 & 45.7\\
MPRD~\cite{ji2021meta}(*) & ICCV'21 & 51.1 & 83.0 & 91.3 & 93.6 & 14.6 & 37.7 & 51.3 & 57.1 \\ 
Xie~\textit{et. al.}~\cite{xie2021unsupervised} & IJMLC'21 & 54.1 & 82.6 & 91.3 & 94.5 & 13.4 & 37.5 & 48.5 & 52.0 \\
DSCE-MC~\cite{yang2021joint} & CVPR'21 & 61.7 & 83.9 & 92.3 & - & 15.5 & 35.2 & 48.3 & - \\
JVTC~\cite{li2020joint} & ECCV'20 & 47.5 & 79.5 & 89.2 & 91.9 & 17.3 & 43.1 & 53.8 & 59.4 \\ 
JGCL~\cite{chen2021joint} & CVPR'21 & 66.8 & 87.3 & 93.5 & 95.5 & 21.3 & 45.7 & 58.6 & 64.5 \\
IICS~\cite{xuan2021intra} & CVPR'21 & 72.9 & 89.5 & 95.2 & 97.0 & 26.9 & 56.4 & 68.8 & 73.4 \\
IIDS~\cite{xuan2022intra} & TPAMI'22 & 78.0 & 91.2 & 96.2 & 97.7 & 35.1 & 64.4 & 76.2 & 80.5\\ 
CAP~\cite{wang2021camera} & AAAI'21 & 79.2 & 91.4 & 96.3 & 97.7 & 36.9 & 67.4 & 78.0 & 81.4\\
CCTSE~\cite{9521886} & TIFS'21 & 67.7 & 89.5  & 94.8 & 96.5 & - & - & - & - \\
CAPL~\cite{liu2023camera} &  NCA'23 & 80.4 & 92.8 & 97.3 & - & 40.7 & 71.2 & 81.4 & - \\
MGH~\cite{wu2021mgh} & ICM'21 & 81.7 & 93.2 & 96.8 & 98.1 & 40.6 & 70.2 & 81.2 & 84.5 \\
ICE~\cite{Chen_2021_ICCV} & ICCV'21 & 82.3 & 93.8 & 97.6 & 98.4 & 38.9 & 70.2 & 80.5 & 84.4 \\
O2CAP~\cite{wang2022offline} & TIP'22 & 82.7 & 92.5 & 96.9 & 98.0 & 42.4 & 72.0 & 81.9 & 85.4 \\
O2CAP-IBN~\cite{wang2022offline} & TIP'22 & 83.7 & 93.1 & 97.4 & 98.1 & \bv{46.9} & \bv{75.5} & \bv{84.8} & \bv{87.7} \\
CASTOR-ICE~\cite{xu2022pseudo} & TITS'22 & 82.8 & 93.6 & 97.5 & 98.5 & 41.7 & 72.3 & 82.3 & 85.8 \\ 
CASTOR-CCL~\cite{xu2022pseudo} & TITS'22 & 84.5 & 93.0 & 97.8 & 98.6 & 33.2 & 61.9 & 74.0 & 78.2 \\ 
Liu~\textit{et. al.}~\cite{liu2022unsupervised} & TIP'22 & 82.4 & 93.0 & - & - & 38.4 & 68.6 & - & - \\
Liu~\textit{et. al.}-IBN~\cite{liu2022unsupervised} & TIP'22 & 82.0 & 92.8 & - & - & 42.4 & 71.6 & - & - \\
CIFL~\cite{pang2022camera} & TMM'22 & 82.4 & 93.9 & 97.9 & 98.1 & 38.8 & 70.1 & 80.7 & 83.9 \\
RTMem~\cite{yin2023real} & TIP'23 & 83.1 & 93.9 & 97.7 & 98.4 & 40.8 & 72.0 & 81.5 & 84.6\\
MCSL~\cite{he2023multiple} & OPTIK'23 & 83.5 & 93.7 & 97.5 & 98.4 & 38.7 & 71.1 & 80.8 & 84.3 \\
PPLR~\cite{cho2022part} & CVPR'22 & 84.4 & 94.3 & 97.8 & 98.6 & 42.2 & 73.3 & 83.5 & 86.5 \\
PPSL~\cite{wu2022pseudo} & TIP'22 & 68.7 & 88.6 & 95.2 & 96.6 & 40.9 & 71.1 & 83.3 & 87.0 \\
PPSL(Concat)~\cite{wu2022pseudo} & TIP'22 & 82.3 & 94.1 & 97.4 & \bv{98.8} & 43.1 & 73.2 & 89.4 & 90.8 \\
PEG~\cite{zhai2022population} & IJCV'22 & 84.5 & 94.3 & 98.0 & 98.5 & 44.9 & 73.9 & 83.2 & 86.3 \\ 
PPCL+CAP~\cite{zheng2022plausible} & TCSVT'22 & 82.4 & 94.0 & 98.1 & - & 37.8 & 70.8 & 80.7 & - \\
PPCL+ICE~\cite{zheng2022plausible} & TCSVT'22 & 82.8 & 93.9 & 97.6 & - & 39.8 & 70.8 & 81.2 & - \\
DiDAL~\cite{10105456} & TMM'23 & 84.8 & 94.2 & \bv{98.2} & - & 45.4 & 74.0 & 84.3 & - \\
CCL~\cite{zhang2023camera} & TCSVT'23 & 85.3 & 94.1 & 97.8 & \bv{98.8} & 41.8 & 71.4 & - & - \\ \hline
\multicolumn{10}{|c|}{\textit{Multi-part based models}}  \\ \hline
PPLR~\cite{cho2022part} & CVPR'22 & 81.5 & 92.8 & 97.1 & 98.1 & 31.4 & 61.1 & 73.4 & 77.8 \\
LPur~\cite{lan2023learning} & TIP'23 & \bv{85.8} & \bv{94.5} & 97.8 & 98.7 & 39.5 & 67.9 & 78.0 & 81.6 \\ \hline
%UPRSTS & ArXiV & \bv{82.4} & \gv{93.0} & \gv{97.5} & - & \bv{72.2} & \bv{84.9} & \bv{92.3} & - & \gv{38.4} & \gv{68.6} & \gv{79.4} & - \\ \hline
%\textbf{Ours} & \textbf{This work} & \rv{82.2} & \rv{92.3} & \rv{96.5} & \rv{97.5} & \gv{70.3} & \gv{83.6} & \rv{90.2} & \gv{92.2} & \rv{37.5} & 63.5 & 73.7 & \rv{77.4} \\ 
\multicolumn{10}{|c|}{\textit{Tracklet-based}}  \\ \hline
Star-Dac~\cite{prasad2022spatio} & PR'21 & 33.9 & 67.0 & 80.6 & 84.9 & - & - & - & - \\
%TAUDL~\cite{li2018unsupervised} & ECCV'18 & 41.2 & 63.7 & - & - & 43.5 & 61.7 & - & - & - & - & - & - \\
TSSL~\cite{wu2020tracklet} & AAAI'20 & 43.3 & 71.2 & - & - & - & - & - & - \\ 
UTAL~\cite{li2019unsupervised} & TPAMI'20 & 46.2 & 69.2 & - & - & 13.1 & 31.4 & - & - \\
CycAs~\cite{wang2020cycas} & ECCV'20 & 64.8 & 84.8 & - & - & 26.7 & 50.1 & - & - \\ 
UGA~\cite{wu2019unsupervised} & ICCV'19 & 70.3 & 87.2 & - & - & 21.7 & 49.5 & - & - \\ \hline
\multicolumn{10}{|c|}{\textit{Fully-Unsupervised}}  \\ \hline
\textbf{Ours} & & \bv{85.8} & 94.0 & 97.7 & 98.5 & 43.2 & 70.9 & 80.8 & 84.2 \\ \hline
%\textbf{Ours} & \textbf{This work} & \bv{82.2} & \bv{92.3} & \bv{96.5} & \bv{97.5} & \bv{70.3} & \bv{83.6} & \bv{90.2} & \bv{92.2} & \bv{37.5} & \bv{63.5} & \bv{73.7} & \bv{77.4} \\ 
\end{tabular}
% *MPRD originally does not use camera information in their pipeline. However, it leverages CamStyle~\cite{zhong2018camstyle} to perform augmentation, \\ which is based on camera labels.
\end{table*}

Even though our model does not use camera labels, we still have competitive performance, and the best mAP for \texttt{Market}. Considering the tracklet-based models, we outperform them in all metrics. Despite the tracklets being a meta-information about the pedestrian motion, which enables the possibility to leverage temporal information, our method better mines the discriminant information just relying upon people's still images.

\begin{table}[ht]
\caption{Comparison with relevant Vehicle ReID methods in the \texttt{Veri776} dataset considering some meta-information or camera/viewpoint labeling. The best results are highlighted in \bv{blue}.}
\label{tab:state_of_art_vehicle}
\centering
\begin{tabular}{|p{2.2cm}|p{1.5cm}|p{0.5cm}|p{0.5cm}|p{0.5cm}|}
\hline
\multicolumn{1}{|c|}{} &
\multicolumn{1}{|c|}{} &
\multicolumn{3}{|c|}{Veri} \\
\hline
Method & Reference & mAP & R1 & R5 \\ \hline
\multicolumn{5}{|c|}{\textit{Camera/Viewpoint-based}} \\ \hline
SSL~\cite{lin2020unsupervised} & CVPR'20 & 23.8 & 69.3 & 72.1 \\
VAPC~\cite{zheng2021aware} & TITS'21 & 30.4 & 76.2 & 81.2 \\
CAPL~\cite{liu2023camera} &  NCA'23 & 41.1 & 87.3 & 91.3 \\
O2CAP~\cite{wang2022offline} & TIP'22 & 41.9 & 87.5 & 92.7\\
O2CAP-IBN~\cite{wang2022offline} & TIP'22 & 42.4 & 89.6 & \bv{93.5}\\
CCL~\cite{zhang2023camera} & TCSVT'23 & 42.6 & 87.0 & -\\
Liu~\textit{et. al.}~\cite{liu2022unsupervised} & TIP'22 & 43.2 & 87.0 & -\\
Liu~\textit{et. al.}-IBN~\cite{liu2022unsupervised} & TIP'22 & 43.9 & 88.9 & -\\
PPLR~\cite{cho2022part} & CVPR'22 & 43.5 & 88.3 & 92.7 \\ 
DiDAL~\cite{10105456} & TMM'23 & 43.5 & \bv{89.0} & 93.5 \\
CTACL~\cite{yu2022camera} & ICRA'22 & 44.2 & 81.6 & 89.5 \\
%CTACL-DA~\cite{yu2022camera} & ICRA'22 & \bv{55.2} & \bv{89.3} & \bv{93.9}\\\hline 
\hline
\multicolumn{5}{|c|}{\textit{Segmentation-based}} \\ \hline
MAPLD~\cite{lu2023mask} & TITS'23 & 33.4 & 78.7 & 83.5\\ \hline
\multicolumn{5}{|c|}{\textit{Multi-part-based models}} \\ \hline
PPLR~\cite{cho2022part} & CVPR'22 & 41.6 & 85.6 & 91.1\\ 
MLPL~\cite{he2022multi} & TVT'22 & \bv{45.1} & 88.3 & 91.1\\ \hline
\multicolumn{5}{|c|}{\textit{Attribute-based Models}} \\ \hline
Method & Reference & mAP & R1 & R5\\ \hline
VRPRD~\cite{bashir2019vr} & PR'19 & 40.1 & 83.2 & 91.1 \\ \hline
\multicolumn{5}{|c|}{\textit{Fully-Unsupervised}} \\ \hline
\textbf{Ours} & & 41.3 & 86.3 & 89.9\\
\hline
\end{tabular}
\end{table}

Similar conclusions can be reached for the \texttt{Veri} dataset, as shown in Table~\ref{tab:state_of_art_vehicle}. Considering mAP, the best method is MLPL~\cite{he2022multi} which relies on multi-part analysis that improves feature description but adds complexity to the training process. Our method relies solely on the feature map extracted from the bounding boxes, without any kind of sub-part analysis. Other methods employ camera or viewpoint labeling, making the task easier. Indeed, the best R1 is achieved by DiDAL~\cite{10105456}, which employs camera labels for optimization. Some methods employ segmentation (MAPLD~\cite{lu2023mask}) or color information (VRPRD~\cite{bashir2019vr}) to help optimization, but our model performs better than both without any kind of supervision or side information. 

% Similar analysis we have for Vehicles in the Veri dataset in Table~\ref{tab:state_of_art_vehicle}. The best method in mAP is MLPL~\cite{he2022multi} which relies on multi-part analysis to improve the feature description, which more complexity to the training process, while ours and other methods rely in the whole image feature map without any kind of sub-part analysis. However, other methods employ the camera or viewpoint labeling, which allows their models to effectively improve the mAP values. Indeed, the best R1 is from DiDAL~\cite{10105456} which employs the camera labeling as part of their training. Other methods still employ segmentation (MAPLD~\cite{lu2023mask}) or color information (VRPRD~\cite{bashir2019vr}) to help model training but our model is better than both in those categories without any kind of supervision or side information. 

\begin{table}[ht]
\caption{Comparison with relevant fully-unsupervised Vehicle ReID methods in \texttt{VehicleID}. The best result is highlighted in \bv{blue}.}
\label{tab:vehicleID}
\centering
\begin{tabular}{|p{1.2cm}| p{1.0cm}|p{0.7cm}|p{0.4cm}|p{0.4cm}|p{0.4cm}|p{0.4cm}|p{0.4cm}|}
\hline
\multicolumn{2}{|c|}{} &
\multicolumn{2}{|c|}{\textbf{TS  = 800}} &
\multicolumn{2}{|c|}{\textbf{TS  = 1600}} &
\multicolumn{2}{|c|}{\textbf{TS  = 2400}} \\
\hline
%\multicolumn{14}{|c|}{\textit{Camera-based}} \\ \hline
%\multicolumn{14}{|c|}{\textit{Segmentation-based}} \\ \hline
\multicolumn{8}{|c|}{\textit{Part-based Models}} \\ \hline
Method & Reference & mAP & R1 & mAP & R1 & mAP & R1\\ \hline
MLPL~\cite{he2022multi} & TVT'22 & \bv{65.3} & \bv{61.1} & \bv{62.7} & \bv{57.3} & \bv{59.6} & \bv{52.4} \\ \hline
\multicolumn{8}{|c|}{\textit{Fully Unsupervised}} \\ \hline
Method & Reference & mAP & R1 & mAP & R1 & mAP & R1\\ \hline
\textbf{Ours} & & 61.7 & 56.0 & 59.7 & 53.4 & 56.9 & 50.1\\
\hline
\end{tabular}
\end{table}

\begin{table*}[ht]
\caption{Comparison with relevant Vehicle ReID methods in \texttt{Veri-Wild} dataset considering some meta-information or camera/viewpoint labeling. The best result is highlighted in \bv{blue}.}
\label{tab:veri_wild}
\centering
\begin{tabular}{|p{1.3cm}| p{1.0cm}|P{0.4cm}|P{0.4cm}|P{0.4cm}|P{0.4cm}|P{0.4cm}|P{0.4cm}|P{0.4cm}|P{0.4cm}|P{0.4cm}|}
\hline
\multicolumn{1}{|c|}{} &
\multicolumn{1}{|c|}{} &
\multicolumn{3}{|c|}{\textbf{Veri-Wild (Small)}} &
\multicolumn{3}{|c|}{\textbf{Veri-Wild (Medium)}} &
\multicolumn{3}{|c|}{\textbf{Veri-Wild (Large)}} \\
\hline
Method & Reference & mAP & R1 & R5 & mAP & R1 & R5 & mAP & R1 & R5 \\ \hline
\multicolumn{11}{|c|}{\textit{Camera/Viewpoint-based}} \\ \hline
SSL~\cite{lin2020unsupervised} & CVPR'20 & 16.1 & 38.5 & 58.1 & 17.9 & 36.4 & 56.0 & 13.6 & 32.7 & 48.2\\ 
VAPC~\cite{zheng2021aware} & TITS'21 & 33.0 & \bv{72.1} & \bv{87.7} & 28.1 & 64.3 & 83.0 & 22.6 & 55.9 & 75.9\\
CTACL~\cite{yu2022camera} & ICRA'22 & \bv{58.2} & 71.1 & 86.6 & \bv{49.2} & \bv{69.2} & 83.7 & \bv{41.2} & \bv{60.1} & \bv{81.5}\\ \hline
%CTACL-DA~\cite{yu2022camera} & ICRA'22 & 65.0 & 79.2 & 93.6 & 56.2 & 73.1 & 89.5 & 44.9 & 63.6 & 83.5\\\hline
\multicolumn{11}{|c|}{\textit{Segmentation-based}} \\ \hline
MAPLD~\cite{lu2023mask} & TITS'23 & 36.6 & \bv{72.1} & 87.6 & 33.4 & 66.2 & \bv{84.5} & 27.7 & 55.9 & 77.3\\ \hline
\multicolumn{11}{|c|}{\textit{Fully-Unsupervised}} \\ \hline
\textbf{Ours} & & 29.9 & 54.1 & 76.6 & 25.8 & 46.4 & 69.2 & 20.0 & 36.2 & 59.0\\
%\textbf{Ours(OSNet)} & & \gv{18.6} & 35.8 & \ov{59.3} & \bv{70.8} & \ov{15.5} & 29.1 & 50.5 & \bv{62.8} & 11.2 & 21.5 & 40.0 & \bv{50.3} \\
%\textbf{Ours(DenseNet121)} & & 14.6 & 27.5 & 50.9 & 63.4 & 12.1 & 23.4 & 42.3 & 54.3 & 8.4 & 16.7 & 31.4 & 41.7\\
%\textbf{Ours(Ensemble)} & & \ov{17.7} & 31.7 & 56.4 & \gv{69.6} & 14.5 & 25.9 & 47.4 & \gv{60.4} & 10.2 & 18.7 & 35.8 & \gv{47.1}\\
\hline
\end{tabular}
\end{table*}

In \texttt{Veri-Wild} (Table~\ref{tab:veri_wild}), following previous conclusions, camera and viewpoint labeling provides a strong capacity to improve model learning. Differently than in \texttt{Veri}, the segmentation-based model MAPLD~\cite{lu2023mask} achieves higher performance than ours. Since \texttt{Veri-Wild} is the most challenging dataset, any side information has the potential to aid model learning. As our model operates in the fully-unsupervised scenario, it sometimes faces performance drops compared to other methods that rely upon meta-information or labeling.


\bibliographystyle{IEEEtran}
\bibliography{IEEEabrv, refs}